\documentclass[final]{cvpr}

\usepackage{times}
\usepackage{epsfig}
\usepackage{graphicx}
\usepackage{amsmath}
\usepackage{amssymb}

\usepackage{siunitx}
\usepackage{cite}
\usepackage{ctable}
\usepackage{hhline}
\usepackage{booktabs}
\usepackage{csquotes}
\usepackage{pifont}
\usepackage{multirow}
\usepackage{booktabs}
\usepackage{capt-of}
\usepackage{subcaption}
\usepackage{makecell}
\usepackage[super]{nth}

\newcolumntype{L}[1]{>{\raggedright\let\newline\\\arraybackslash\hspace{0pt}}m{#1}}
\newcolumntype{C}[1]{>{\centering\let\newline\\\arraybackslash\hspace{0pt}}m{#1}}
\newcolumntype{R}[1]{>{\raggedleft\let\newline\\\arraybackslash\hspace{0pt}}m{#1}}

\newcommand{\cmark}{\ding{51}}%
\newcommand{\xmark}{\ding{55}}%
\newcommand{\quotes}[1]{``#1''}

\def\eg{\emph{e.g.}}

\makeatletter
\def\blfootnote{\xdef\@thefnmark{}\@footnotetext}
\makeatother

\usepackage[pagebackref=true,breaklinks=true,colorlinks,bookmarks=false]{hyperref}

\def\cvprPaperID{****} % *** Enter the CVPR Paper ID here
\def\confYear{CVPR 2021}
\begin{document}

%%%%%%%%% TITLE
\title{IntegralAction: Pose-driven Feature Integration for \\Robust Human Action Recognition in Videos}

\author{
Gyeongsik Moon$^{*1}$\hspace{1.0cm} Heeseung Kwon$^{*2}$\hspace{1.0cm} Kyoung Mu Lee$^1$\hspace{1.0cm} Minsu Cho$^2$\\
\\
\hspace{0.5cm}$^{1}$SNU ECE \& ASRI \hspace{2.5cm}
$^{2}$POSTECH CSE \& AIGS\\
{\small \texttt {\{mks0601, kyoungmu\}@snu.ac.kr} \hspace{1.0cm} \texttt {\{aruno, mscho\}@postech.ac.kr}}
}

\maketitle

\blfootnote{* equal contribution}

%%%%%%%%% ABSTRACT
\begin{abstract}
Most current action recognition methods heavily rely on appearance information by taking an RGB sequence of entire image regions as input. While being effective in exploiting contextual information around humans, e.g., human appearance and scene category, they are easily fooled by out-of-context action videos where the contexts do not exactly match with target actions. In contrast, pose-based methods, which take a sequence of human skeletons only as input, suffer from inaccurate pose estimation or ambiguity of human pose per se. Integrating these two approaches has turned out to be non-trivial; training a model with both appearance and pose ends up with a strong bias towards appearance and does not generalize well to unseen videos. To address this problem, we propose to learn pose-driven feature integration that dynamically combines appearance and pose streams by observing pose features on the fly. The main idea is to let the pose stream decide how much and which appearance information is used in integration based on whether the given pose information is reliable or not. We show that the proposed IntegralAction achieves highly robust performance across in-context and out-of-context action video datasets. The codes are available in \href{https://github.com/mks0601/IntegralAction_RELEASE}{here}.
\end{abstract}

\section{Introduction}

Human action recognition in videos aims at classifying an input video of human action into one of pre-defined target classes~\cite{carreira2017quo,xie2018rethinking,feichtenhofer2019slowfast,lin2019tsm,simonyan2014two,feichtenhofer2016convolutional,yan2018spatial,li2019actional,choutas2018potion,yan2019pa3d}.
Following the success of convolutional neural networks (CNNs) on image classification~\cite{simonyan2014very,he2016deep}, video action recognition has made remarkable progress by developing deep neural models that process RGB image frames via spatio-temporal convolution~\cite{carreira2017quo,xie2018rethinking,feichtenhofer2019slowfast,lin2019tsm} or two-stream convolution with scene optical flow~\cite{simonyan2014two,feichtenhofer2016convolutional}. These {\em appearance-based} methods, however, learn to exploit contextual information (\textit{e.g.},  scene class, dominant objects, or background motion), rather than human action performed in the video~\cite{li2018resound,li2019repair}. This is a critical issue in terms of robustness since they are all vulnerable to the attack of out-of-context actions, \eg, mimes~\cite{weinzaepfel2019mimetics}.   
While recent {\em pose-based} methods~\cite{yan2018spatial,li2019actional,choutas2018potion,yan2019pa3d}, which replace the RGB input with human skeletons, have a potential to resolve this issue as a reasonable alternative, the situation is very difficult in the wild. For most real-world videos in the standard benchmark datasets, human poses are not easy to detect, only partially visible (mostly, close-up faces and hands), or completely absent (first-person view without any persons). Even with successful detection, human poses are often ambiguous without context.

\begin{figure*}
\begin{center}
\includegraphics[width=0.8\linewidth]{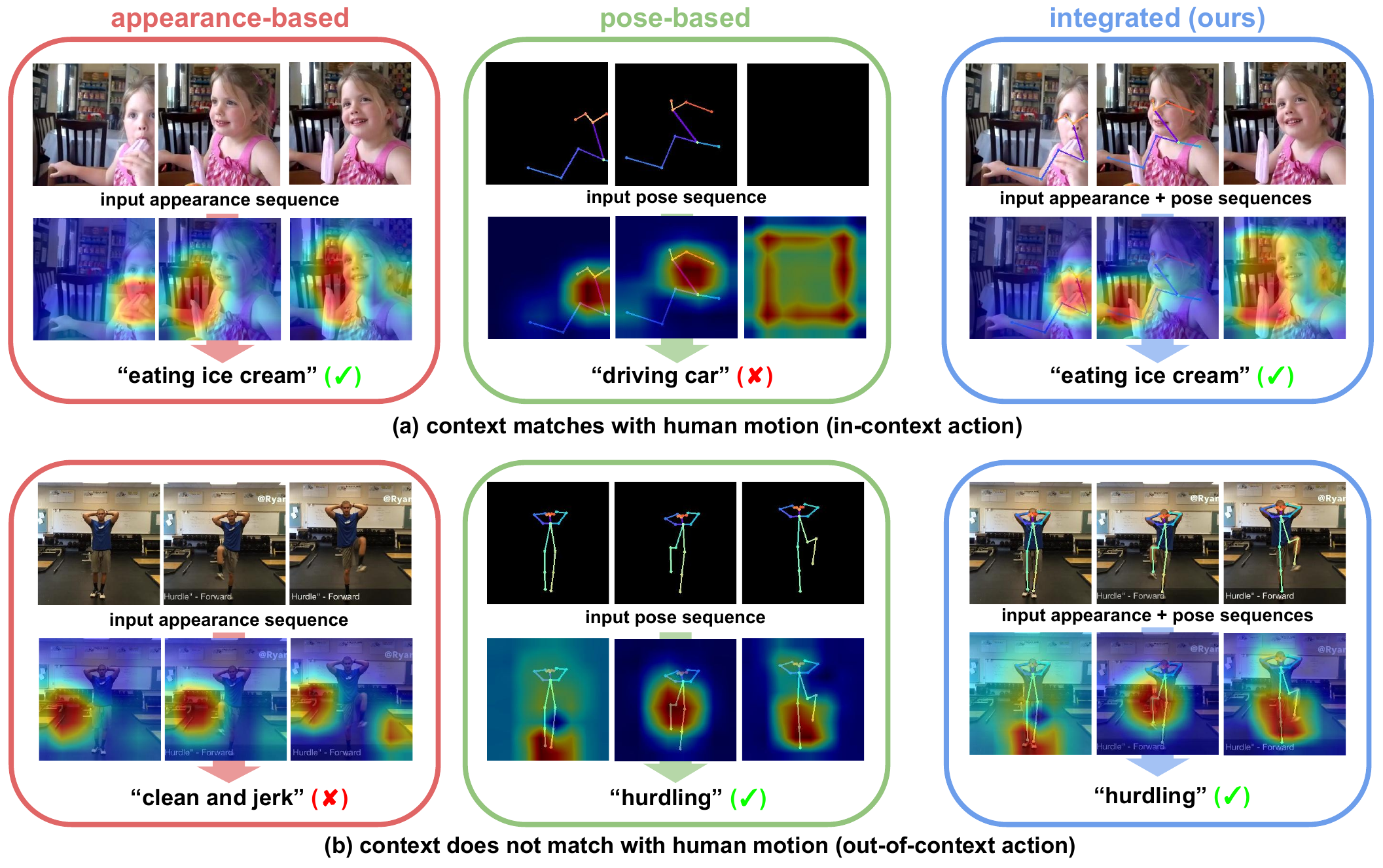}
\end{center}
\vspace*{-7mm}
  \caption{
  Comparative examples of appearance-based, pose-based, vs. our approaches. For (a) in-context action video and (b) out-of-context action video cases, representative frames and their class activation maps~\cite{zhou2016learning} are visualized with the action prediction results. The appearance-based and pose-based approaches are ours with only one stream without the integration. In the input pose sequence, the blank image indicates a pose estimation failure.
  }
\vspace*{-4mm}
\label{fig:intro_compare_appearance_pose}
\end{figure*}
As shown in the example in Fig.~\ref{fig:intro_compare_appearance_pose}(a), 
while the appearance-based approach successfully predicts \quotes{eating ice cream} by exploiting the appearance of ice creams, the pose-based approach often fails due to the ambiguity of the pose without the appearance context. 
In contrast, the pose-based approach has the strong advantage of robustness in understanding actual human actions. As shown in the example of Fig.~\ref{fig:intro_compare_appearance_pose}(b), while the appearance approach is misled by the appearance of barbels, the pose-based approach correctly predicts ``hurdling''. 
Several methods~\cite{choutas2018potion,yan2019pa3d,du2017rpan,wang2018pose,yan2018spatial,zolfaghari2017chained,luvizon20182d} attempt to integrate the two approaches, but the problem has turned out to be non-trivial. 
 They either aggregates the two streams by averaging predicted action scores~\cite{choutas2018potion,yan2019pa3d,du2017rpan,wang2018pose,yan2018spatial} or fusing features from each stream by fully-connected layers~\cite{zolfaghari2017chained,luvizon20182d}.
 Their models all inherit a strong bias of contextual information present in the datasets~\cite{li2018resound,li2019repair} and thus are easily fooled by out-of-context action videos~\cite{weinzaepfel2019mimetics}.

To address the problem, we propose an effective and robust integration model, dubbed {\em IntegralAction}, that dynamically combines the appearance and pose information in a pose-driven manner.
The main idea of pose-driven feature integration is to let the pose stream decide how much and which appearance information is used in integration based on whether the given pose information is reliable or not. 
It thus encourages the system to filter out unnecessary contextual information and focus on human motion information when the pose information is sufficient for action recognition.
In experimental evaluations, we demonstrate that the proposed pose-driven feature integration greatly improves action recognition on out-of-context action videos, Mimetics~\cite{weinzaepfel2019mimetics} dataset, without losing performance on in-context action videos, Kinetics~\cite{kay2017kinetics} and NTU-RGBD~\cite{shahroudy2016ntu} datasets.

\section{Related work}

\noindent \textbf{Appearance-based human action recognition.}
The appearance-based human action recognition methods rely on raw RGB video frames for prediction.
There exist different approaches to it using Recurrent Neural Networks (RNNs) \cite{donahue2015long}, two-stream CNNs \cite{simonyan2014two,feichtenhofer2016convolutional}, and 3D CNNs \cite{tran2015learning,carreira2017quo,tran2018closer,xie2018rethinking}.
Two-stream approaches~\cite{simonyan2014two,feichtenhofer2016convolutional} adopt two-stream networks with RGB and optical flow streams, which process RGB frames together with corresponding optical flow frames.
Carreira~\etal~\cite{carreira2017quo} introduce the inflated 3D CNN (I3D) that expands ImageNet~\cite{russakovsky2015imagenet} pre-trained kernels of 2D CNN to 3D.
They verify 3D CNNs with large-scale in-context action datasets such as Kinetics~\cite{kay2017kinetics} become strong baselines in action recognition.
Du~\etal~\cite{tran2018closer} and Xie~\etal~\cite{xie2018rethinking} factorize 3D convolutions into 2D and 1D convolutions for efficient prediction.
Recently, several methods propose innovative architectures for efficient action recognition~\cite{lin2019tsm,feichtenhofer2019slowfast,feichtenhofer2020x3d}.
Lin~\etal~\cite{lin2019tsm} propose the temporal shifting module (TSM) that enables them to learn spatio-temporal features using 2D convolutions only.
Feichtenhofer~\etal~\cite{feichtenhofer2019slowfast,feichtenhofer2020x3d} propose SlowFast networks, which capture spatial semantics and motion separately by applying different spatio-temporal resolutions for two different networks.
They also propose an efficient set of X3D architectures~\cite{feichtenhofer2020x3d} by controlling scaling factors such as channel widths, layer depths, frame intervals, and spatio-temporal resolutions for action recognition.

\noindent
\textbf{Pose-based human action recognition.}
The pose-based human action recognition methods use human skeleton or pose videos for prediction.
The pose video can be obtained from depth sensors~\cite{liu2016spatio,keselman2017intel} or off-the-shelf 2D multi-person pose estimation models~\cite{cao2017realtime,he2017mask}.
The earlier work of Jhuang~\etal~\cite{jhuang2013towards} shows that human pose features can improve human action recognition in videos.
Cheron~\etal~\cite{cheron2015p} construct two-stream networks that process RGB frames and optical flows of human parts.
Choutas~\etal~\cite{choutas2018potion} introduce a concept of pose motion that represents the temporal movement of human pose in a spatial domain.
Yan~\etal~\cite{yan2019pa3d} propose a unified 3D CNN model that effectively encodes multiple pose modalities.
Du~\etal~\cite{du2017rpan} and Rohit~\etal~\cite{girdhar2017attentional} use estimated human poses for attention mechanisms in their model by converting human keypoints to 2-dimensional heatmaps.
Recently, GraphCNN-based methods have been proposed for skeleton-based action recognition~\cite{yan2018spatial,shi2019two,li2019actional,cheng2020skeleton,liu2020disentangling}.
Yan~\etal~\cite{yan2018spatial} construct a spatio-temporal human pose graph and process it by GraphCNN.
Shi~\etal~\cite{shi2019two} introduce a GraphCNN-based action recognition system that learns graph topology.
Li~\etal~\cite{li2019actional} propose an encoder-decoder structure to capture action-specific latent high-order dependencies in their GraphCNN-based model.
Cheng~\etal~\cite{cheng2020skeleton} propose a shift graph operation for reducing FLOPs of GraphCNNs.
Liu~\etal~\cite{liu2020disentangling} propose a novel graph convolution operator (G3D) for capturing long-range joint relationship modeling.

\noindent 
\textbf{Integrating appearance and pose.}
Several methods have attempted to improve performance in action recognition by integrating the two approaches~\cite{choutas2018potion,yan2019pa3d,du2017rpan,girdhar2017attentional,zolfaghari2017chained,wang2018pose,yan2018spatial,luvizon20182d}.
Most of the methods~\cite{choutas2018potion,yan2019pa3d,du2017rpan,wang2018pose,yan2018spatial} use to simply add predicted action scores from the two models in their testing stage.
Rohit~\etal~\cite{girdhar2017attentional} propose to use a bilinear pooling between appearance and pose features for their action classifier.
Zolfaghari~\etal~\cite{zolfaghari2017chained} develop a multi-stream 3D CNN for processing multi-frame human poses, optical flows, and RGBs together.
Luvizon~\etal~\cite{luvizon20182d} propose a multi-task system that simultaneously predicts 2D/3D human pose and action class.

Compared to our method, all these integration methods have serious limitations in robustness to in-context or out-of-context actions.
Their models heavily rely on the appearance stream without any gating or regularization so that they are not able to filter out a strong contextual bias from out-of-context action videos. 
Furthermore, the post-processing integration of~\cite{choutas2018potion,yan2019pa3d,du2017rpan,wang2018pose,yan2018spatial} is easily affected by inaccurate action prediction from the pose stream. 
Our dynamic integration method of IntegralAction effectively suppresses misleading contextual features from the appearance stream and also supplements inaccurate or ambiguous pose information with useful contextual appearance.

\noindent \textbf{Gating for neural networks.} 
Gating mechanisms control the information flow of neural networks through a multiplicative interaction and are being widely used for a variety of tasks~\cite{dauphin2017language, hu2018squeeze, miech2017learnable, xie2018rethinking}. 
Dauphin~\etal~\cite{dauphin2017language} use a simple gating block for language modeling.
Hu~\etal~\cite{hu2018squeeze} introduce a squeeze-and-excitation block for image classification.
For video classification, Miech~\etal~\cite{miech2017learnable} and Xie~\etal~\cite{xie2018rethinking} propose gating modules to reassign the distribution of feature channels. 
In this paper, we propose a new gating mechanism, which controls the information flow of appearance and human pose information through dynamic integration for recognizing human actions.

Our contributions can be summarized as follows.
\begin{itemize}
\item We present an action recognition model, IntegralAction, that dynamically integrates appearance and pose information in a pose-driven way.

\item We show that our pose-driven feature integration significantly improves the action recognition accuracy on out-of-context action videos while preserving the accuracy on in-context action videos.

\item Our IntegralAction achieves highly robust performance across in-context and out-of-context action video datasets. Especially, it significantly outperforms the recent state-of-the-art methods on out-of-context action video dataset.
\end{itemize}

\section{IntegralAction model}
\begin{figure}[t]
\begin{center}
\includegraphics[width=1.0\linewidth]{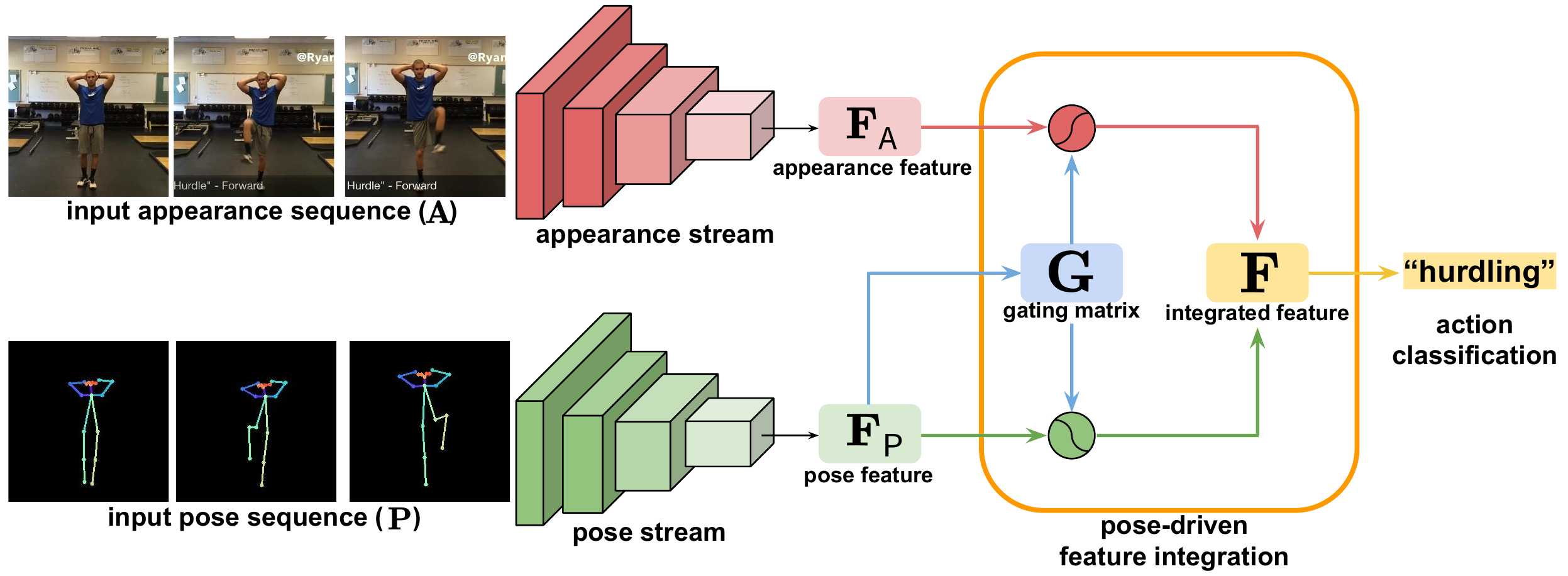}
\end{center}
\vspace*{-7mm}
   \caption{
The overall pipeline of the proposed IntegralAction.
   }
\vspace*{-5mm}
\label{fig:overall_pipeline}
\end{figure}

Given a video clip, the proposed model, IntegralAction, takes as input both its appearance sequence $\mathbf{A}$ and the corresponding pose estimation sequence $\mathbf{P}$, and then predicts action probabilities of the video over target classes $\mathcal{C}$. 
The appearance sequence is composed of RGB frames, while the pose sequence is of human pose frames. 
As illustrated in Fig.~\ref{fig:overall_pipeline}, IntegralAction processes the two sequences, $\mathbf{A}$ and $\mathbf{P}$, via appearance and pose streams;
these two streams transform $\mathbf{A}$ and $\mathbf{P}$ into spatio-temporal appearance and pose features, respectively. 
The proposed pose-driven feature integrator then combines the two features into an integrated action feature.
Finally, the action feature is fed to a classifier, which outputs action probabilities over target classes $\mathcal{C}$.

\subsection{Appearance stream}

For an input to the appearance stream, we sample $T$ frames from the given video clip and construct a sequence $\mathbf{A} \in \mathbb{R}^{T \times 3 \times H_\text{A} \times W_\text{A}}$ where $T$, $H_\text{A}$, and $W_\text{A}$ denote the numbers of frames, height, and width of the sequence, respectively. 
The sampled frames are selected by randomly choosing the first frame within the video clip and then collecting subsequent frames at $\tau$ frame intervals~\cite{carreira2017quo,xie2018rethinking}. The frames are resized and cropped into the size of $H_\text{A} \times W_\text{A}$. 
Given the sequence of RGB frames $\mathbf{A}$, the appearance stream transforms it into an appearance feature $\mathbf{F}_\text{A} \in \mathbb{R}^{T \times C}$.

For an architecture of the stream, we employ ResNet~\cite{he2016deep} and add the temporal shift module (TSM)~\cite{lin2019tsm} for each residual block of the ResNet for efficient and effective spatio-temporal feature extraction following Lin~\etal~\cite{lin2019tsm}; 
the TSM enables to obtain the effect of 3D convolutions using efficient 2D convolutions by shifting a part of input feature channels along the temporal axis before the convolution operation.
Following the setting in~\cite{lin2019tsm}, we shift $1/8$ of the input feature channels forward and another $1/8$ of the channels backward in the TSM.
The final appearance feature $\mathbf{F}_\text{A}$ is obtained by performing a global average pool on the output of the ResNet.

\subsection{Pose stream}

For an input to the pose stream, we sample $T$ frames from the video clip using the same sampling scheme used in the appearance stream,  and extract human body keypoints from the frames using off-the-shelf 2D multi-person pose estimation methods~\cite{cao2017realtime,he2017mask,cao2018openpose}.
The keypoints for each frame are then translated into $K$ keypoint heatmaps~\cite{cao2018openpose,xiao2018simple} and $B$ part affinity fields~\cite{cao2017realtime}, which will be described below.  
Combining the heatmaps and fields along channels, we construct a sequence $\mathbf{P} \in \mathbb{R}^{T \times (K+B) \times H_\text{P} \times W_\text{P}}$ where $H_\text{P}$ and $W_\text{P}$ represent the height and width of the sequence, respectively.
Given the sequence of human pose frames $\mathbf{P}$, the pose stream transforms it into a pose features $\mathbf{F}_\text{P} \in \mathbb{R}^{T \times C}$.

For an architecture of the stream, we employ the same ResNet model used in the appearance stream but with some modification for memory efficiency. Specifically, we skip the first two-strided convolutional layer and the next max-pooling layer of the ResNet; this allows us to set $H_\text{P}$ and $W_\text{P}$ to one-quarter of $H_\text{A}$ and $W_\text{A}$, respectively while preserving the spatial size of the output to be the same as that from the appearance stream. As the channel dimension of $\mathbf{P}$, $K+B$, is relatively large, we reduce the spatial size of $\mathbf{P}$ for efficient feature extraction.
To adapt the channel dimension of $\mathbf{P}$ to the input channel dimension of the first block of the ResNet, we put a front-end convolutional block that consists of a 3-by-3 convolutional layer, batch normalization~\cite{ioffe2015batch}, and ReLU function.

\smallbreak
\noindent \textbf{Keypoint heatmap.}
Each keypoint is translated into a heatmap $\mathbf{H}_k$ using a Gaussian blob~\cite{cao2018openpose,xiao2018simple}:
$\mathbf{H}_k(x,y) =  \exp{\left(-\frac{(x-x_k)^2+(y-y_k)^2}{2\sigma^2}\right)},$
where $x_k$ and $y_k$ are the coordinates of $k$th keypoint and $\sigma$ is set to 0.5. 
When multiple persons are detected in the frame, we sort them by the pose estimation score and select top-5 persons.
If the score is below 0.1, the keypoint heatmap of the person becomes zero heatmap.
Then, we accumulate heatmaps of each keypoint from the persons and clamp the value to a maximum of 1.0.
\smallbreak
\noindent \textbf{Part affinity field (PAF).}
The PAF~\cite{cao2017realtime} is a vector field between human keypoint locations in the image space.
 For its construction, we define a bone as a pair of parent and child nodes in the human skeleton graph. To translate the $b$th bone of a  detected human into a PAF $\mathbf{L}_b$, we assign the 2D orientation vector of the corresponding keypoint pair to all positions $(x,y)$ on the line formed by the keypoint pair: $\mathbf{L}_b(x,y) = \mathbf{v}_b$ where $\mathbf{v}_b \in \mathbb{R}^2$ is the 2D unit vector in the direction of the $b$th bone~\cite{cao2017realtime}. 
 Like the keypoint heatmap, we sort pose estimation results by the score and select top-5 persons to make PAFs.
 The PAF of a person whose score is lower than 0.1 becomes zero PAF.
 When a position is assigned bones of multiple persons, it obtains the average of all the corresponding vectors. 
 PAFs implicitly provide information about which person each keypoint belongs to, \eg, greedy parsing of PAFs would be able to reconstruct the keypoint connections (\textit{i.e.}, bones) of a person. PAFs are thus useful in the presence of multiple persons and complementary to the keypoint heatmaps.

\subsection{Pose-driven feature integration}
\begin{figure}[t]
\begin{center}
\includegraphics[width=1.0\linewidth]{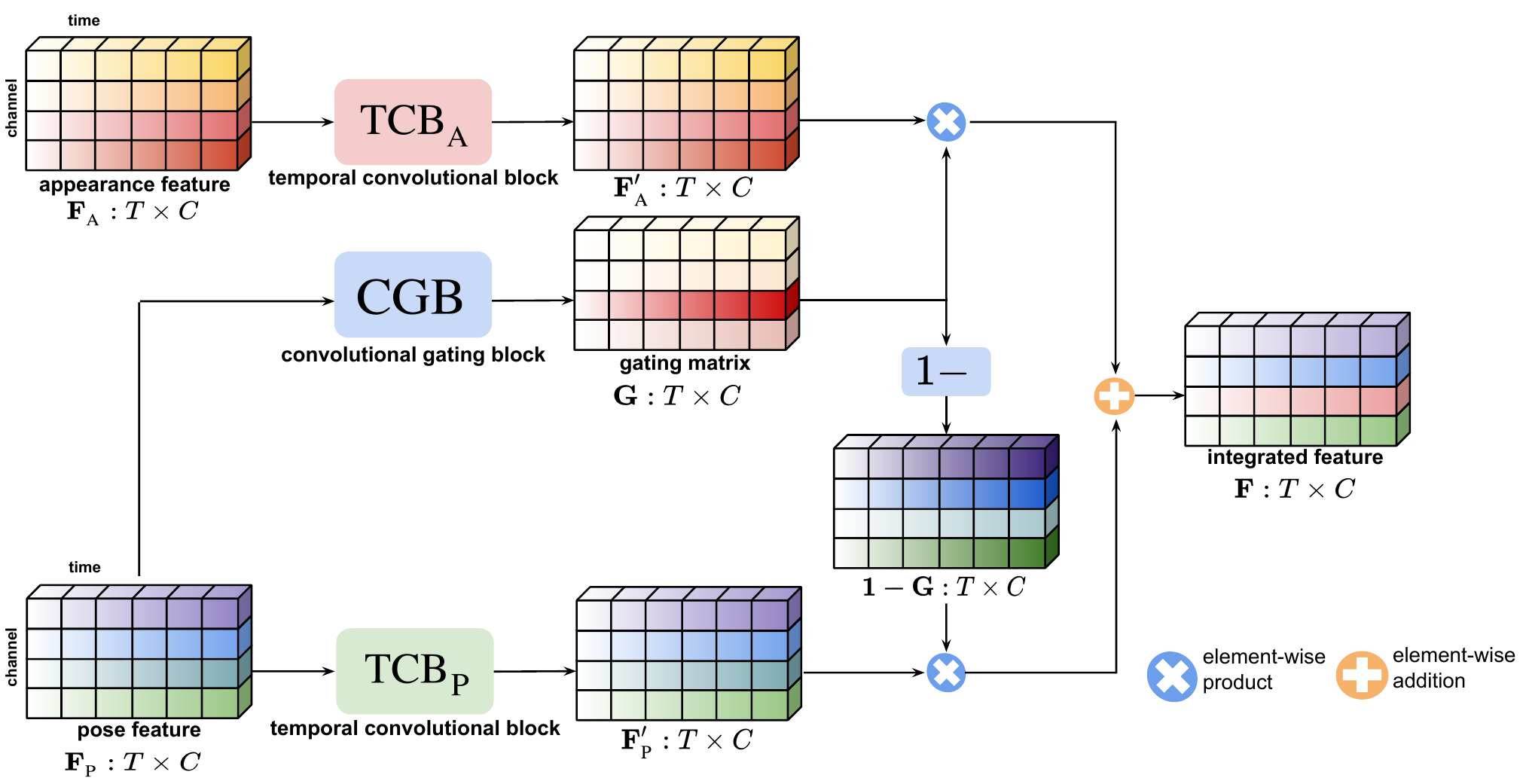}
\end{center}
\vspace*{-7mm}
   \caption{
Pose-driven feature integration.
   }
\vspace*{-5mm}
\label{fig:gate_architecture}
\end{figure}
To combine the appearance and pose features, $\mathbf{F}_\text{A}$ and $\mathbf{F}_\text{P}$, into an {\em adaptive} and {\em robust} action feature $\mathbf{F}$, we propose to learn {\em pose-driven feature integration}. 
As illustrated in Fig.~\ref{fig:gate_architecture}, it proceeds in three steps: (1) feature alignment, (2) pose-driven gating, and (3) aggregation. 

\smallbreak
\noindent \textbf{Feature alignment.} 
The appearance and pose features, $\mathbf{F}_\text{A}$ and $\mathbf{F}_\text{P}$, are transformed 
into $\mathbf{F}_\text{A}'$ and $\mathbf{F}_\text{P}'$ so that they are semantically aligned (\textit{i.e.}, channel-wise aligned) to each other with the same channel dimension. 
This is done by two temporal convolution blocks, TCB$_\text{A}$ and TCB$_\text{P}$, each of which consists of convolution with kernel size $1$, layer normalization~\cite{ba2016layer}, and ReLU activation function; the two temporal convolution blocks are separately applied to $\mathbf{F}_\text{A}$ and $\mathbf{F}_\text{P}$, respectively, and produce outputs with the same channel dimension as inputs. 
This semantic alignment along channels is necessary for the two features, $\mathbf{F}_\text{A}'$ and $\mathbf{F}_\text{P}'$, to be integrated by addition after gating. 

\smallbreak
\noindent \textbf{Pose-driven gating.} The pose-driven gating takes as input the pose feature $\mathbf{F}_\text{P}$, predicts a channel-wise gating matrix $\mathbf{G} \in \mathbb{R}^{T \times C}$, and imposes it on semantically aligned features $\mathbf{F}_\text{A}'$ and $\mathbf{F}_\text{P}'$. 
The gating matrix $\mathbf{G}$ is predicted by a convolutional gating block, CGB, that consists of convolution with kernel size 1, batch normalization, and sigmoid activation function.
The gating matrix $\mathbf{G}$, where each element lies $[ 0, 1]$, is used to perform gating on $\mathbf{F}_\text{A}'$ and $\mathbf{F}_\text{P}'$ in opposite directions:
$\mathbf{G} \odot \mathbf{F}_{\text{A}}'$
and  
$(1-\mathbf{G}) \odot \mathbf{F}_{\text{P}'}$, 
where $\odot$ denotes element-wise multiplication.

\smallbreak
\noindent \textbf{Aggregation.}
Finally, the aggregation simply combines the gated features of appearance and pose by element-wise addition, which produces an integrated action feature $\mathbf{F} \in \mathbb{R}^{T \times C}$ as follows:
\begin{equation}
\mathbf{F} =  \mathbf{G} \odot \mathbf{F}_{\text{A}}'
+ (1-\mathbf{G}) \odot \mathbf{F}_{\text{P}}'.
\end{equation}

 The main idea behind the pose-driven feature integration is to let the pose feature decide how much and which appearance information is used in integration based on whether the given pose information is reliable or not. To this end, we enforce it to learn a {\em priority for pose} by training the gating module with a regularizer: 
 \begin{equation}
 L_{\text{gate}} = -\log (1-\mathbf{G}), 
 \end{equation}
which encourages the gate to be open to the appearance feature only when the cost needs to be taken based on the given pose feature. This prevents the integration from being biased towards strong contextual features from the appearance stream. 
If the pose feature provides sufficient information for action recognition, the proposed integration will focus on it without the risk of being misled by strong contexts from the appearance feature.
Otherwise, it will utilize more contextual information from the appearance feature.

\subsection{Action classification}

The final classifier uses the integrated action feature $\mathbf{F}$ to predict probabilities over target action classes. The classification is performed for each frame of $\mathbf{F}$ by a fully-connected layer with softmax output. 
As in~\cite{wang2016temporal,lin2019tsm}, we average the probabilities of the frames to obtain the final action probabilities of the video. 

We train our model, IntegralAction, by minimizing the loss function: 
 \begin{equation}
 L = L_\text{cls} + \lambda L_{\text{gate}},
 \end{equation}
where $L_\text{cls}$ represents the standard cross entropy loss and  $L_\text{gate}$ is the gate regularizer described above. $\lambda$ is a balancing factor between them. 

The training procedure proceeds as follows. We first train an appearance-only model and a pose-only model, which use $\mathbf{F}_\text A$ and $\mathbf{F}_\text P$ for classification, respectively, by minimizing $L_{\text{cls}}$ only. 
These pre-trained appearance-only and pose-only models are used to initialize the IntegralAction except for the parts of feature integration and action classification, which are randomly initialized. The feature integration and classification parts are then trained by minimizing the entire loss $L$ while the pre-trained parts of the network are frozen.

\subsection{Implementation details}

PyTorch~\cite{paszke2017automatic} is used for implementation. 
The ResNet is initialized with the publicly released weight pre-trained on the ImageNet~\cite{russakovsky2015imagenet}, and the weights of the remaining part are initialized by Gaussian distribution with $\sigma=0.001$. 
The weights are updated by the SGD with a mini-batch size of 32.
$C=512$ is used for all experiments.
We set the size of the input RGB sequence as $224 \times 224$, and that of the input pose sequence as $56 \times 56$.
We pre-train the appearance and pose streams with the initial learning rate of $10^{-2}$.
The learning rate is reduced by a factor of 10 at \nth{30} and \nth{60} epoch until \nth{70} epoch when trained on Kinetics~\cite{kay2017kinetics} and reduced by a factor of 10 at \nth{20} and \nth{30} epoch until \nth{40} epoch when trained on Kinetics50~\cite{kay2017kinetics} and NTU-RGBD~\cite{shahroudy2016ntu}. 
After the pre-training, we continue the training of full IntegralAction.
To this end, the initial learning rate is set to $10^{-3}$ and reduced by a factor of 10 at \nth{10} and \nth{15} epoch until \nth{20} epoch.
Random scaling, translation, and horizontal flip augmentations are applied during the training following Lin~\etal~\cite{lin2019tsm}.
We use four NVIDIA RTX 2080Ti GPUs for training our IntegralAction. 
In testing time, we sampled 10 clips per video and averaged the prediction following Lin~\etal~\cite{lin2019tsm}.

\section{Experiments}

\subsection{Datasets and evaluation metrics}
\noindent \textbf{Kinetics.}
Kinetics~\cite{kay2017kinetics} is a large-scale human action dataset mostly with in-context action videos.
The dataset consists of 240K training and 20K validation videos with 400 action classes.
Following the work of~\cite{weinzaepfel2019mimetics}, we also use a subset of Kinetics, Kinetics50, which contains 33K training and 2K validation videos with 50 action classes; these 50 action classes are strongly related to human body motions rather than background scene and objects.
For human pose estimation on Kinetics, we use the results released by Yan~\etal~\cite{yan2018spatial}, which are obtained using OpenPose~\cite{cao2018openpose}.

\noindent \textbf{Mimetics.}
Mimetics~\cite{weinzaepfel2019mimetics} is a human action dataset mostly with out-of-context action videos.
It consists of 713 videos and 50 human action classes, which are the same as those of Kinetics50.
Due to its small scale, it is \emph{only used to evaluate models trained on Kinetics50 or Kinetics} following Weinzaepfel~\etal~\cite{weinzaepfel2019mimetics}.
For human pose estimation on Mimetics, we use ResNet50-based Mask R-CNN~\cite{he2017mask,wu2019detectron2}.

\noindent \textbf{NTU-RGBD.}
NTU-RGBD~\cite{shahroudy2016ntu} contains 56K RGBD video clips captured from a controlled lab environment.
40 subjects perform 60 actions, and most of the videos correspond to in-context actions.
We use the standard cross-subject split for training and testing and exploit 2D human pose annotations provided by~\cite{shahroudy2016ntu}.

Following the previous results reported on Kinetics, Mimetics~\cite{carreira2017quo,weinzaepfel2019mimetics}, and NTU-RGBD~\cite{yan2018spatial,weinzaepfel2019mimetics,shi2019two,li2019actional}, both top-1 and top-5 accuracies are used as evaluation metrics for Kinetics and Mimetics while top-1 accuracy is used as an evaluation metric for NTU-RGBD.

\begin{table*}
\footnotesize
\centering
\setlength\tabcolsep{1.0pt}
\def\arraystretch{1.1}
\caption{Top-1 and top-5 accuracy comparison between baselines and the proposed method on Kinetics, Mimetics, and NTU-RGBD.}
\vspace*{-4mm}
\label{table:ours}
\begin{tabular}{C{5.5cm}|C{0.9cm}C{0.9cm}|C{0.9cm}C{0.9cm}|C{0.9cm}C{0.9cm}||C{0.9cm}}
\specialrule{.1em}{.05em}{.05em}
\multirow{ 2}{*}{methods} & \multicolumn{2}{c|}{Kinetics} & \multicolumn{2}{c|}{Mimetics} & \multicolumn{2}{c||}{NTU-RGBD} & avg\\
& top-1 & top-5 & top-1  & top-5  & top-1  & top-5 & rank \\ \hline
appearance-only model & \textbf{73.5} & \textbf{91.0} & 6.3 & 16.7 & 90.4 & 99.0 & 2.7 \\
pose-only model & 30.2 & 51.3 & 15.2 & \textbf{33.1} & 83.7 & 97.0 & 3.2\\
\hline
\textbf{IntegralAction (ours, $\lambda=1.5$)}  & 73.3 & 90.8 & 12.8 & 26.0  & \textbf{91.7} & \textbf{99.4} & \textbf{1.8}\\
\textbf{IntegralAction (ours, $\lambda=5.0$)}  & 65.0 & 85.9 & \textbf{15.3} & 31.5  & 91.0 & 99.2 & 2.2\\
 \specialrule{.1em}{.05em}{.05em}
\end{tabular}
\vspace*{-5mm}
\end{table*}

\subsection{Results on the benchmark datasets}
In this experiment, we use ResNet-50 for the appearance-only model and use ResNet-18 for the pose-only model.
We set $T=8$ and $\tau=8$ for the appearance-only model, and $T=32$ and $\tau=2$ for the pose-only model. 
The same settings are used for appearance and pose streams of our integrated model.
We apply $4\times1\times1$ average pooling along the temporal axis for $\mathbf{F}_\text P$ to set the same feature dimension as that of $\mathbf{F}_\text A$ before the feature integration.

\noindent
\textbf{Comparison to the baselines.}
First of all, we compare single-stream models with the proposed integrated model on different benchmark datasets. Since the balancing factor $\lambda$ controls the degree of pose priority in dynamic gating, we use two versions of IntegralAction: weak priority ($\lambda$ = 1.5) and strong priority ($\lambda$ = 5.0).
The results are summarized in Table~\ref{table:ours}. 
The appearance-only model shows strong performance on the in-context action datasets (Kinetics and NTU-RGBD) but drastically fails on the out-of-context action dataset (Mimetics). The pose-only model shows the opposite; it achieves strong performance on Mimetics while significantly underperforming others on Kinetics and NTU-RGBD.
The proposed method, IntegralAction, presents robust performance on all the benchmark datasets.
It achieves good performance on Mimetics while being comparable to the appearance-only model on Kinetics.
In NTU-RGBD, our method obtains an additional gain by integrating appearance and pose information.
The last column of Table~\ref{table:ours} represents the average rank over all the three datasets, which shows that the proposed method performs the best on average.

\begin{table}
\footnotesize
\centering
\caption{Top-1 and top-5 accuracy comparison with state-of-the-art methods on Kinetics and Mimetics. Methods with * initialize weights with ImageNet pre-trained ones.}
\vspace*{-3mm}
\label{table:sota_kinetics}
\begin{tabular}{C{4.1cm}|C{0.6cm}C{0.6cm}|C{0.6cm}C{0.6cm}}
\specialrule{.1em}{.05em}{.05em}
\multirow{ 2}{*}{methods} & \multicolumn{2}{c|}{Kinetics} & \multicolumn{2}{c}{Mimetics} \\
& top-1  & top-5  & top-1  & top-5  \\ \hline
\textbf{\textit{Appearance-based methods}} & & &  &   \\
I3D (RGB)*~\cite{carreira2017quo} & 71.1 & 89.3 & - & -  \\
R(2+1)D (RGB)*~\cite{tran2018closer} & 72.0 & 90.0 & - & -  \\
TSM*~\cite{lin2019tsm} & 73.5 & 91.0 & 6.3 & 16.7  \\
3D ResNext-101 (RGB)*~\cite{weinzaepfel2019mimetics} & 74.5 & - & 8.6 & 20.1  \\
3D ResNext-101 (two-stream)*~\cite{weinzaepfel2019mimetics} & - & - & 10.5 &  26.9 \\ 
SlowFast Res-101~\cite{feichtenhofer2019slowfast} & 78.9 & 93.5 & - & -  \\
X3D-XL~\cite{feichtenhofer2020x3d} & \textbf{79.1} & \textbf{93.9} &- &- \\
\hline
\textbf{\textit{Pose-based methods}} &  &  &  & \\
Deep LSTM~\cite{shahroudy2016ntu} & 16.4 & 35.3 & - & - \\
TCN~\cite{kim2017interpretable} & 20.3 & 40.0 & - & -\\
ST-GCN~\cite{yan2018spatial} & 30.7 & 52.8 & 12.6 &  27.4 \\
SIP-Net~\cite{weinzaepfel2019mimetics} & 32.8 & - & 14.2 &  \textbf{32.0} \\
\hline
\textbf{IntegralAction (ours, $\lambda=1.5$)*}  & 73.3 & 90.8 & 12.8 & 26.0 \\
\textbf{IntegralAction (ours, $\lambda=5.0$)*} & 65.0 & 85.9 & \textbf{15.3} & 31.5  \\
 \specialrule{.1em}{.05em}{.05em}
\end{tabular}
\vspace*{-4mm}
\end{table}

\begin{table}
\footnotesize
\centering
\caption{Top-1 accuracy comparison with state-of-the-art methods on NTU-RGBD.}
\vspace*{-3mm}
\label{table:sota_NTU}
\begin{tabular}{C{4.0cm}|C{1.5cm}}
\specialrule{.1em}{.05em}{.05em}
methods & accuracy \\ \hline
Lie Group~\cite{vemulapalli2014human} & 50.1 \\
Du~\etal~\cite{du2015hierarchical} & 59.1 \\
Deep LSTM~\cite{shahroudy2016ntu} & 60.7 \\
SIP-Net~\cite{weinzaepfel2019mimetics} & 64.8 \\
Zolfaghari~\etal~\cite{zolfaghari2017chained} & 67.8 \\
TCN~\cite{kim2017interpretable} & 74.3 \\
ST-GCN~\cite{yan2018spatial} & 81.5 \\
AS-GCN~\cite{li2019actional} & 86.8 \\
2S-AGCN~\cite{shi2019two} & 88.5 \\
Shift-GCN~\cite{cheng2020skeleton} & 90.7 \\
MS-G3DNet~\cite{liu2020disentangling} & 91.5 \\
\hline
\textbf{IntegralAction (ours, $\lambda=1.5$)} & \textbf{91.7} \\
\textbf{IntegralAction (ours, $\lambda=5.0$)} & 91.0 \\
 \specialrule{.1em}{.05em}{.05em}
\end{tabular}
\vspace*{-3mm}
\end{table}

\noindent
\textbf{Comparison to the state-of-the-art methods.}
We compare our results with recent state-of-the-art results on Kinetics and Mimetics in Table~\ref{table:sota_kinetics}. 
Each section of the table contains the results of appearance-based methods~\cite{carreira2017quo,tran2018closer,lin2019tsm,weinzaepfel2019mimetics}, pose-based methods~\cite{shahroudy2016ntu,kim2017interpretable,yan2018spatial,weinzaepfel2019mimetics}, and the proposed method, respectively.
As expected, appearance-based methods are better than pose-based methods on Kinetics, while pose-based methods are better than appearance-based methods on Mimetics.
The proposed method is competitive with the appearance-based methods~\cite{carreira2017quo,tran2018closer,lin2019tsm,weinzaepfel2019mimetics}, and outperforms all the other pose-based methods~\cite{shahroudy2016ntu,kim2017interpretable,yan2018spatial,weinzaepfel2019mimetics} by a large margin in Kinetics.
At the same time, our method achieves state-of-the-art performance in Mimetics.
Table~\ref{table:sota_NTU} summarizes the comparative results on NTU-RGBD. 
The upper part of the table contains the results of previous pose-based methods including a hand-crafted feature method~\cite{vemulapalli2014human}, RNN methods~\cite{du2015hierarchical,shahroudy2016ntu}, CNN methods~\cite{zolfaghari2017chained,kim2017interpretable,weinzaepfel2019mimetics}, and GraphCNN methods~\cite{yan2018spatial,li2019actional,shi2019two}.
The proposed method outperforms all the other methods, outperforming the state of the arts by the margin of 3.2\% points at top-1 accuracy.

\subsection{Comparison with other integration methods}
We compare top-1 and top-5 accuracy from appearance-only model, pose-only model, and various feature integration methods including feature fuse~\cite{zolfaghari2017chained,luvizon20182d} and score average in testing stage~\cite{choutas2018potion,yan2019pa3d,du2017rpan,wang2018pose,yan2018spatial} in Table~\ref{table:ablation_agg}.
We additionally train variants of our IntegralAction by disabling the gating or gating without $L_{\text{gate}}$.
Among all the integration methods, our setting achieves the best accuracy on Mimetics and marginally lower accuracy than the best performing one on Kinetics50.
The relative performance difference between other best performing integration methods and ours on Mimetics is \emph{23\%}, while that on Kinetics50 is \emph{3\%}.
For all models, the network architectures are based on that of our IntegralAction.
The appearance and pose streams are based on ResNet-18, and we set $T=\tau=8$.
All models are trained on Kinetics50 and then tested on Kinetics50 and Mimetics.

\begin{table}
\footnotesize
\centering
\setlength\tabcolsep{1.0pt}
\def\arraystretch{1.1}
\caption{Top-1 and top-5 accuracy comparison between various appearance and pose stream integration methods on Kinetics50 and Mimetics. The numbers in parentheses represent relative difference between other best performing integration method and ours.}
\vspace*{-3mm}
\label{table:ablation_agg}
\begin{tabular}{C{3.0cm}|C{1.2cm}C{1.2cm}|C{1.2cm}C{1.2cm}}
\specialrule{.1em}{.05em}{.05em}
\multirow{ 2}{*}{methods} & \multicolumn{2}{c|}{Kinetics50} & \multicolumn{2}{c}{Mimetics} \\
& top-1 & top-5 & top-1  & top-5  \\ \hline
appearance-only model & 72.8 & 91.7 & 11.2 & 31.7  \\
pose-only model & 45.6 & 72.9 & 26.0 &  \textbf{52.2} \\ \hline
feature fuse~\cite{zolfaghari2017chained,luvizon20182d} & 73.8 &  \textbf{93.4} & 19.5 & 42.3  \\
score average~\cite{choutas2018potion,yan2019pa3d} & 73.9 & 91.4 & 21.6 &  48.7 \\
ours without gating  & \textbf{74.2} & 93.2 & 21.3 & 45.5 \\
ours without $L_{\text{gate}}$ & 73.2 & 92.1 & 15.9 & 37.2  \\
\textbf{IntegralAction (ours)} & \shortstack[c]{72.2\\($\downarrow$ 3\%)} & \shortstack[c]{92.3\\($\downarrow$ 1\%)} & \shortstack[c]{\textbf{26.5}\\\textbf{($\uparrow$ 23\%)}} & \shortstack[c]{50.5\\($\uparrow$ 6\%)} \\
 \specialrule{.1em}{.05em}{.05em}
\end{tabular}
\vspace*{-3mm}
\end{table}

\begin{figure}
\begin{center}
   \includegraphics[width=0.7\linewidth]{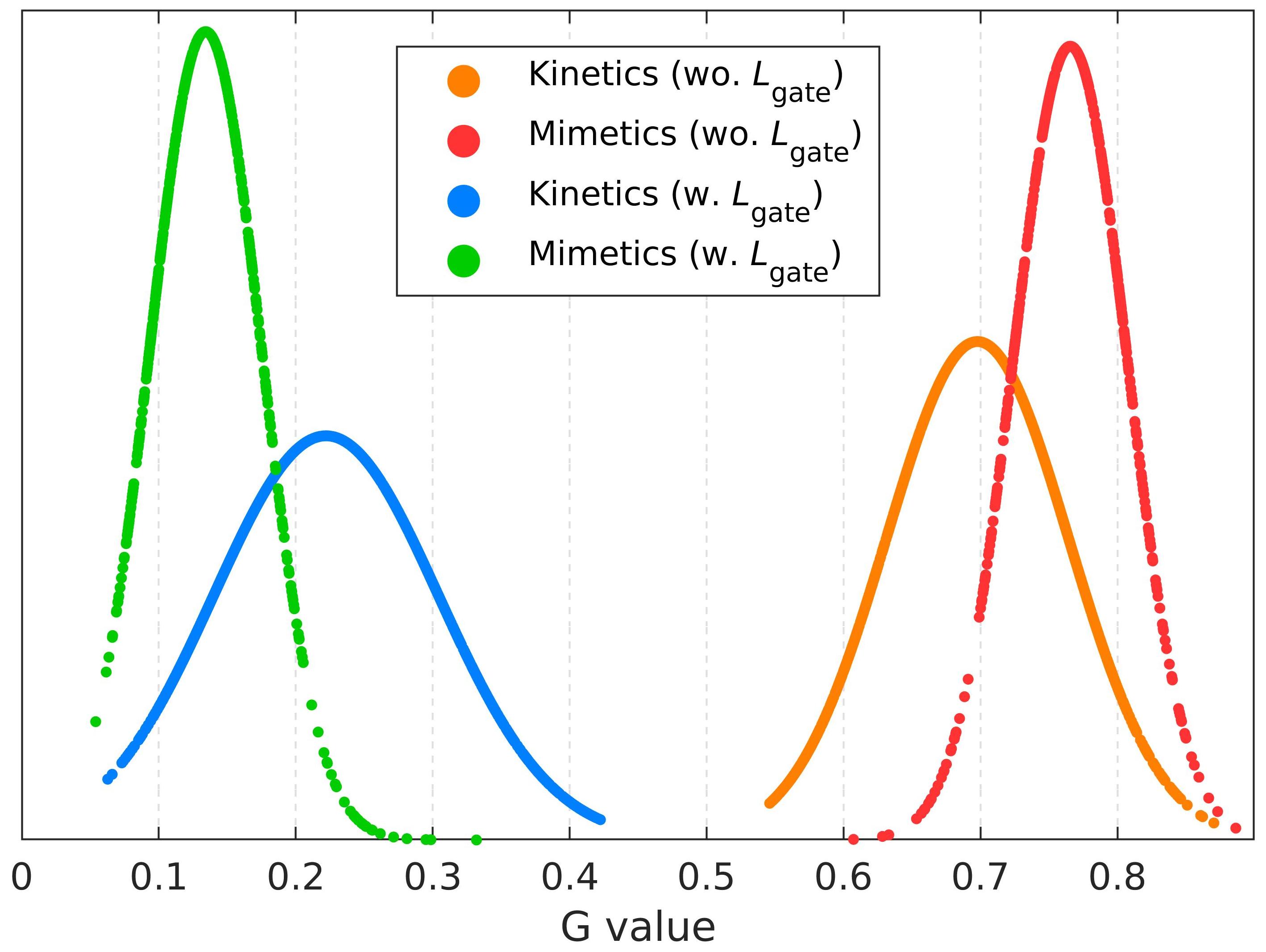}
\end{center}
\vspace*{-7mm}
   \caption{The Gaussian distributions of averaged $\mathbf{G}$ from models trained with and without $L_{\text{gate}}$ on Kinetics50 and Mimetics.}
\vspace*{-3mm}
\label{fig:gate_distribution}
\end{figure}

\noindent\textbf{Comparison with the feature fuse.}
The feature fuse method~\cite{zolfaghari2017chained,luvizon20182d} suffers from low accuracy on Mimetics because their networks are trained without any gating or regularization; therefore, theirs cannot filter out a strong contextual bias from out-of-context action videos.
In contrast, our IntegralAction is not dominated by the biased contextual information by introducing the regularizer $L_\text{gate}$, which results in significant performance improvement on Mimetics as shown in \nth{6} and the last row of the Table~\ref{table:ablation_agg}.
To further validate this, we show the Gaussian distribution of the gating matrix $\mathbf{G}$ averaged over the channel and temporal dimensions from models trained with and without $L_\text{gate}$ on each frame of Kinetics50 and Mimetics in Figure~\ref{fig:gate_distribution}.
As the figure shows, the regularizer translates the mean of each distribution to a lower value, which enforces our system to prefer pose features over the appearance features.
Thus, our IntegralAction is optimized to utilize the pose feature when the input pose sequence provides sufficient information for the action recognition without being dominated by the appearance feature.

To implement the feature fuse, we pass $\mathbf{F}_\text{P}$ and $\mathbf{F}_\text{A}$ to an additional fully-connected layer, of which output is used for the action prediction.

\noindent\textbf{Comparison with the score average.}
The score averaging methods~\cite{choutas2018potion,yan2019pa3d,du2017rpan,wang2018pose,yan2018spatial} average the predicted action probability from the appearance-only and pose-only models in the testing stage.
This makes their system suffer from low accuracy on Mimetics because strongly biased action prediction from the appearance-based model is not filtered out when averaging.
In contrast, our IntegralAction dynamically integrates the appearance features and pose features.
Figure~\ref{fig:gate_distribution} shows that distributions of the averaged gating matrix $\mathbf{G}$ from Kinetics50 and Mimetics have very different variances, which indicates that the estimated gating matrix varies according to the input sequence.
As some videos of Kinetics50 contain human-central videos while others contain context-central videos, whether the input pose sequence is sufficient for the action recognition or not varies a lot, which makes the gating matrix diverse.
In contrast, most of the videos of Mimetics contain human-central videos, which results in a less diverse gating matrix.
We additionally show the gating matrix $\mathbf{G}$ averaged over the channel dimension and all frames of an action class in Table~\ref{table:gate_per_action_class}.
The table shows that the averaged gating matrix becomes lower when the action class is highly related to human motion and becomes higher when the class is related to the context, such as background and objects.
This dynamic feature integration is essential for filtering out the biased action prediction from the appearance feature.

\begin{table}
\footnotesize
\setlength{\tabcolsep}{1pt}
\centering
\caption{Averaged $\mathbf{G}$ values for each action class on Kinetics50.}
\vspace*{-4mm}
\label{table:gate_per_action_class}
\begin{tabular}{C{3.0cm}|C{2.0cm}}
\specialrule{.1em}{.05em}{.05em}
action classes & averaged $\mathbf{G}$ \\ \hline
writing & 0.33  \\
flying kite & 0.32 \\
tying tie & 0.31   \\ 
driving car & 0.28 \\ \hline
hitting baseball & 0.18  \\
skipping rope & 0.17 \\
deadlifting & 0.16 \\
clean and jerk & 0.15 \\
 \specialrule{.1em}{.05em}{.05em}
\end{tabular}
\vspace*{-5mm}
\end{table}

\noindent\textbf{Performance on Kinetics50.}
Although our IntegralAction outperforms other integration methods on Mimetics by a large margin, it decreases the top-1 accuracy of the appearance-only model on Kinetics50 slightly while increases the top-5 accuracy.
We believe this is because our regularizer is applied to all frames equally, which can enforce our model to prefer the pose feature even when the appearance feature plays a critical role in the frames for the action recognition.
We tried to utilize the action recognition test results on the training set to supervise the gating matrix differently for each frame; however, it did not improve the action recognition accuracy in a meaningful margin.
We leave designing a frame (or video)-adaptive regularizer as future work.

\subsection{Ablation study on IntegralAction}
For this ablation study, we use ResNet-18 and set $T=\tau=8$ for both appearance and pose streams.
The model is trained on Kinetics50 and then tested on Kinetics50 and Mimetics.

\noindent \textbf{Pose-driven vs. appearance-driven feature integration.}
To demonstrate the effectiveness of the pose-driven feature integration, we compare the top-1 and top-5 accuracy between models that estimate the gating matrix $\mathbf{G}$ from the appearance feature $\mathbf{F}_\text A$, pose feature $\mathbf{F}_\text P$, and both ones in Table~\ref{table:pose_driven}.
The both-driven method predicts $\mathbf{G}$ from a combined feature, concatenation of $\mathbf{F}_\text A$ and $\mathbf{F}_\text P$ along the channel dimension.
As the table shows, our pose-driven feature integration significantly outperforms appearance-driven and both-driven ones on Mimetics while produces marginally lower accuracy on Kinetics50.
The relative performance gap on Mimetics is \emph{12\%}, while that on Kinetics50 is \emph{2\%}.
The reason for the worse performance of the appearance-driven and both-driven ones on Mimetics is that the appearance feature provides strongly biased contextual information from out-of-context action videos.
Therefore, CGB, which estimates the gating matrix, is dominated by biased contextual information.
In contrast, the pose feature only provides human motion information without contextual information.
This can make CGB robust to the biased context, which leads to more robust performance on out-of-context action videos.

\begin{table}
\footnotesize
\centering
\setlength\tabcolsep{1.0pt}
\def\arraystretch{1.1}
\caption{Top-1 and top-5 accuracy comparison between models that estimate $\mathbf{G}$ from appearance feature, pose feature, and both features on Kinetics50 and Mimetics. The numbers in parentheses represent relative difference between other best performing method and ours.}
\vspace*{-3mm}
\label{table:pose_driven}
\begin{tabular}{C{3.2cm}|C{1.0cm}C{1.0cm}|C{1.0cm}C{1.0cm}}
\specialrule{.1em}{.05em}{.05em}
\multirow{ 2}{*}{methods} & \multicolumn{2}{c|}{Kinetics50} & \multicolumn{2}{c}{Mimetics} \\
& top-1 & top-5 & top-1  & top-5  \\ \hline
appearance-driven & \textbf{73.8} & \textbf{93.0} & 23.2 & 48.2 \\
\textbf{pose-driven (ours)} & \shortstack[c]{72.2\\($\downarrow$ 2\%)} & \shortstack[c]{92.3\\($\downarrow$ 1\%)} & \shortstack[c]{\textbf{26.5}\\\textbf{($\uparrow$ 12\%)}} & \shortstack[c]{\textbf{50.5}\\\textbf{($\uparrow$ 5\%)}} \\
both-driven & 73.3 & 92.4 & 23.7 & 46.5 \\
 \specialrule{.1em}{.05em}{.05em}
\end{tabular}
\vspace*{-3mm}
\end{table}

\begin{table}
\footnotesize
\setlength{\tabcolsep}{1pt}
\centering
\caption{Top-1 and top-5 accuracy comparison between models trained with various $\lambda$ values on Kinetics50 and Mimetics.}
\vspace*{-3mm}
\label{table:ablation_lambda}
\begin{tabular}{C{1.7cm}|C{0.9cm}C{0.9cm}|C{0.9cm}C{0.9cm}}
\specialrule{.1em}{.05em}{.05em}
\multirow{ 2}{*}{$\lambda$ value} & \multicolumn{2}{c|}{Kinetics50} & \multicolumn{2}{c}{Mimetics} \\
& top-1  & top-5  & top-1  & top-5  \\ \hline
0.0 & \textbf{73.2} & 92.1 & 15.9 & 37.2  \\
1.0 & 72.8 & \textbf{92.6} & 25.7 &  47.8 \\
\textbf{1.5 (ours)} & 72.2 & 92.3 & 26.5 & \textbf{50.5} \\
5.0 & 67.8 & 89.7 & \textbf{27.2} & 48.2 \\
 \specialrule{.1em}{.05em}{.05em}
\end{tabular}
\vspace*{-4mm}
\end{table}

\noindent \textbf{Various training settings.}
We provide the top-1 and top-5 accuracy from various $\lambda$ values in Table~\ref{table:ablation_lambda}.
As the table shows, a larger $\lambda$ value regularizes the model to choose the feature from the pose stream stronger, which results in low performance on Kinetics50 and good performance on Mimetics.
We found that $\lambda=1.5$ achieves high accuracy on Mimetics while marginally decreases the accuracy on Kinetics50; therefore, we set $\lambda=1.5$ through the whole experiments.

In addition, we show the top-1 and top-5 accuracy from various possible training strategies of IntegralAction in Table~\ref{table:training_strategy}.
For the comparison, we trained three models, including ours.
The first one is trained without pre-training, and the second one is trained from pre-trained appearance and pose streams, while the two streams are not fixed when training feature integration and classification part.
As the table shows, starting from pre-trained appearance and pose streams greatly improves the accuracy on Mimetics.
In addition, fixing the pre-trained two streams also increases the accuracy on Mimetics significantly.
Therefore, we trained our IntegralAction from pre-trained appearance and pose streams and fixed them during the final training stage.

\begin{table}
\footnotesize
\centering
\caption{Top-1 and top-5 accuracy comparison between models trained with various training strategies on Kinetics50 and Mimetics.}
\vspace*{-3mm}
\label{table:training_strategy}
\begin{tabular}{C{1.4cm}C{0.7cm}|C{0.9cm}C{0.9cm}|C{0.9cm}C{0.9cm}}
\specialrule{.1em}{.05em}{.05em}
\multirow{ 2}{*}{pre-train} & \multirow{ 2}{*}{fix} & \multicolumn{2}{c|}{Kinetics50} & \multicolumn{2}{c}{Mimetics} \\
& & top-1  & top-5 & top-1  & top-5  \\ \hline
\xmark & \xmark  & 72.2 & \textbf{93.6} & 17.8 & 43.1  \\
\cmark & \xmark  & \textbf{72.5} & 92.0 & 23.5 & 46.6  \\
\cmark & \cmark  & 72.2 & 92.3 & \textbf{26.5} & \textbf{50.5}  \\
 \specialrule{.1em}{.05em}{.05em}
\end{tabular}
\vspace*{-7mm}
\end{table}

We think pre-training each stream separately and fixing them can maintain their complementary characteristics better than training the whole system from scratch or fine-tuning them during the final training stage.
Maintaining the complementary characteristics of each stream and utilizing them is the essence of our IntegralAction.

\section{Conclusion}

We propose IntegralAction, which dynamically integrates appearance and pose information in a pose-driven manner for robust human action recognition.
The previous integration methods use a static aggregation of the two information or sequentially refine the pose information using the appearance information, which suffers from a strong bias of contextual information from out-of-context action videos.
In contrast, our pose-driven feature integration can filter out the biased contextual information, thus can perform robust action recognition on both in-context and out-of-context action videos.
The proposed IntegralAction achieves highly robust performance across in-context and out-of-context action video datasets.

\noindent\textbf{Acknowledgments.} 
This work was supported by 
the NRF grant (NRF-2017R1E1A1A01077999 -50\%), the Visual Turing Test project (IITP-2017-0-01780 -50\%), and the IITP grant (No.2019-0-01906, AI Graduate School Program - POSTECH) funded by the Ministry of Science and ICT of Korea.

\clearpage

\begin{center}
\textbf{\large Supplementary Material of \enquote{IntegralAction: Pose-driven Feature Integration for Robust Human Action Recognition in Videos}}
\end{center}

In this supplementary material, we present additional experimental results and studies that are omitted in the main manuscript due to the lack of space.

\section{Effect of the pose stream inputs}
To analyze the effects of keypoint heatmaps and PAFs as inputs of the pose stream, we compare the top-1 and top-5 accuracy from pose-only models that take 1) the keypoint heatmaps, 2) the PAFs, and 3) both in Table~\ref{table:posestream_input}.
The table shows that taking both inputs achieves the best accuracy on Kinetics50 and Mimetics.
The keypoint heatmaps provide the locations of each human body keypoint, which are useful in single person cases, but do not include sufficient information for differentiating each person in multi-person cases.
On the other hand, the PAFs contain relationships between the keypoints from each person, which can provide information to differentiate each person in multi-person cases.
We found that most of the videos in Mimetics contain a single person, which makes the heatmap-only model perform well on the action recognition. However, many videos in Kinetics contain multiple persons, and thus additional PAFs further improve the accuracy.

\section{Deeper comparison with the score averaging}
\begin{figure}
\begin{center}
\includegraphics[width=1.0\linewidth]{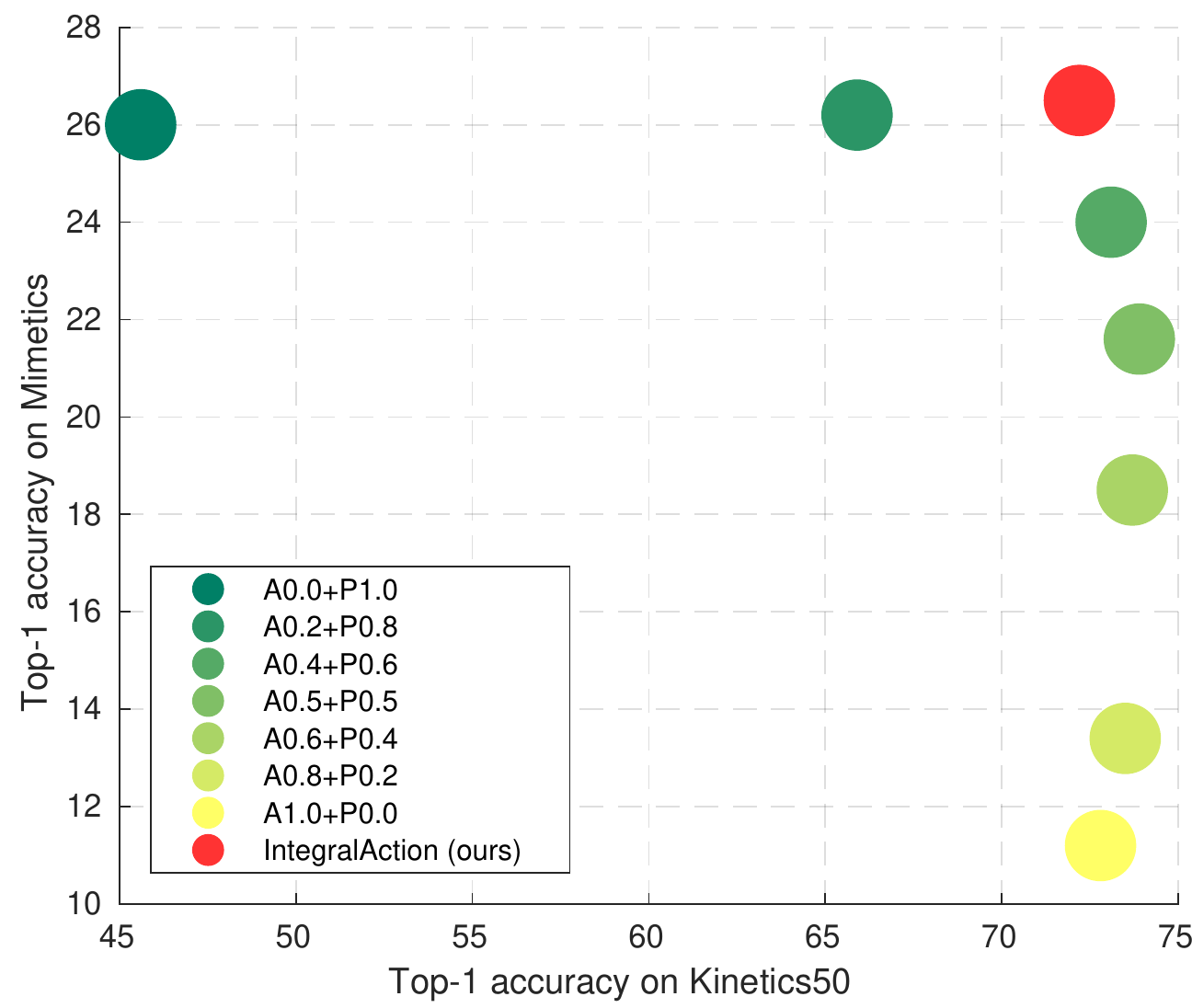}
\end{center}
\vspace*{-3mm}
   \caption{
Top-1 accuracy on Kinetics50 and Mimetics comparison between the proposed IntegralAction and score averaging with various average ratios.
The numbers to the left and right of plus sign denote averaging weight at the score from the appearance-based and pose-based models, respectively.
   }
\vspace*{-3mm}
\label{fig:compare_score_avg}
\end{figure}

Most of the previous methods~\cite{choutas2018potion,yan2019pa3d,du2017rpan,wang2018pose,yan2018spatial} use to simply average predicted action scores from the appearance-based and pose-based action recognition models in their testing stage.
We compared their accuracy with ours in Table 4 of the main manuscript, and we provide a deeper comparison between ours and theirs in Figure~\ref{fig:compare_score_avg}.
We report top-1 accuracy on Kinetics50 and Mimetics of score averaging with various averaging weights.
As the figure shows, the score averaging method that performs best on Kinetics50 achieves slightly better accuracy than ours.
However, it suffers from a noticeable performance drop on Mimetics.
The proposed IntegralAction achieves highly robust performance on both Kinetics50 and Mimetics datasets.

\section{Appearance-only, pose-only, vs. IntegralAction}
\begin{figure*}
\begin{center}
\includegraphics[width=1.0\linewidth]{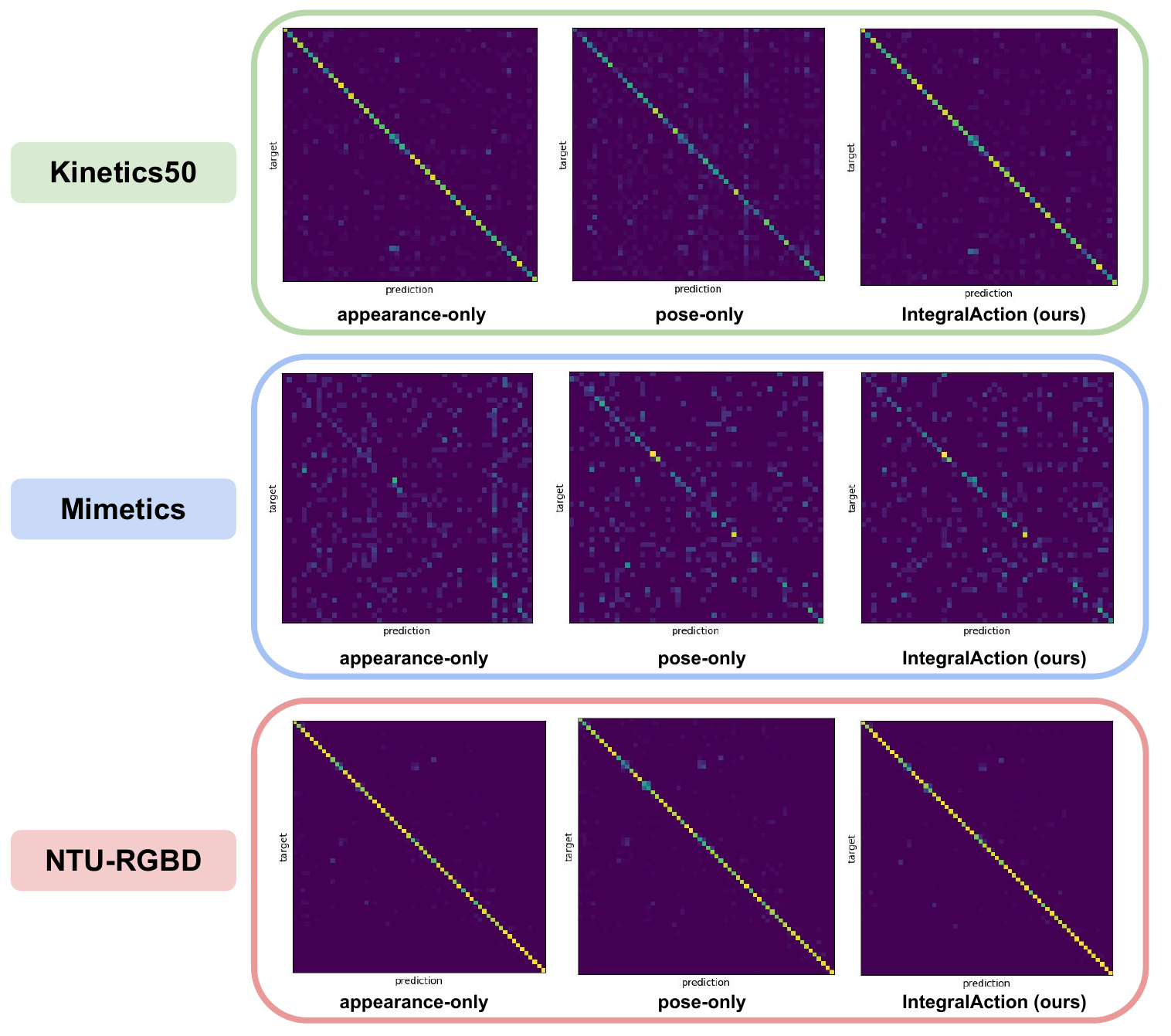}
\end{center}
\vspace*{-7mm}
   \caption{
Visualized confusion matrices of appearance-only, pose-only, and our IntegralAction on Kinetisc50, Mimetics, and NTU-RGBD.
   }
\vspace*{-7mm}
\label{fig:confusion_matrix}
\end{figure*}

In this experiment, we analyze top-1 accuracy of each action class using the appearance-only, pose-only, and the proposed IntegralAction on Kinetics50, Mimetics, and NTU-RGBD in Table~\ref{table:label_accuracy_compare_kinetics50}, ~\ref{table:label_accuracy_compare_mimetics}, and ~\ref{table:label_accuracy_compare_ntu}, respectively. In addition, we compare them with   the oracle selection that chooses the best prediction between the appearance-only and the pose-only.
We also visualize confusion matrices from the appearance-only, pose-only, and our IntegralAction in Fig.~\ref{fig:confusion_matrix}.
As the tables and figures show, our IntegralAction produces robust action recognition over action classes of the three datasets, while the appearance-only and pose-only fail on Mimetics and Kinetics50, respectively.
The proposed IntegralAction achieves the best average accuracy on Mimetics and NTU-RGBD.
In addition, it significantly outperforms the pose-only and achieves comparable average accuracy with the appearance-only on Kinetics50.

\begin{table}
\small
\centering
\setlength\tabcolsep{1.0pt}
\def\arraystretch{1.1}
\caption{Top-1 and top-5 accuracy comparison between pose-only models that take various combinations of input on Kinetics50 and Mimetics.}
\label{table:posestream_input}
\begin{tabular}{C{3.0cm}|C{1.0cm}C{1.0cm}|C{1.0cm}C{1.0cm}}
\specialrule{.1em}{.05em}{.05em}
\multirow{ 2}{*}{settings} & \multicolumn{2}{c|}{Kinetics50} & \multicolumn{2}{c}{Mimetics} \\
& top-1 & top-5 & top-1  & top-5  \\ \hline
heatmap-only & 43.5 & 72.1 & \textbf{26.0} & 50.0 \\
PAF-only & 45.0 & \textbf{73.2} & 25.0 & 49.8 \\
\textbf{heatmap + PAF (ours)} & \textbf{45.6} & 72.9 & \textbf{26.0} & \textbf{52.2}  \\
 \specialrule{.1em}{.05em}{.05em}
\end{tabular}
\end{table}

\begin{figure*}
\begin{center}
\includegraphics[width=1.0\linewidth]{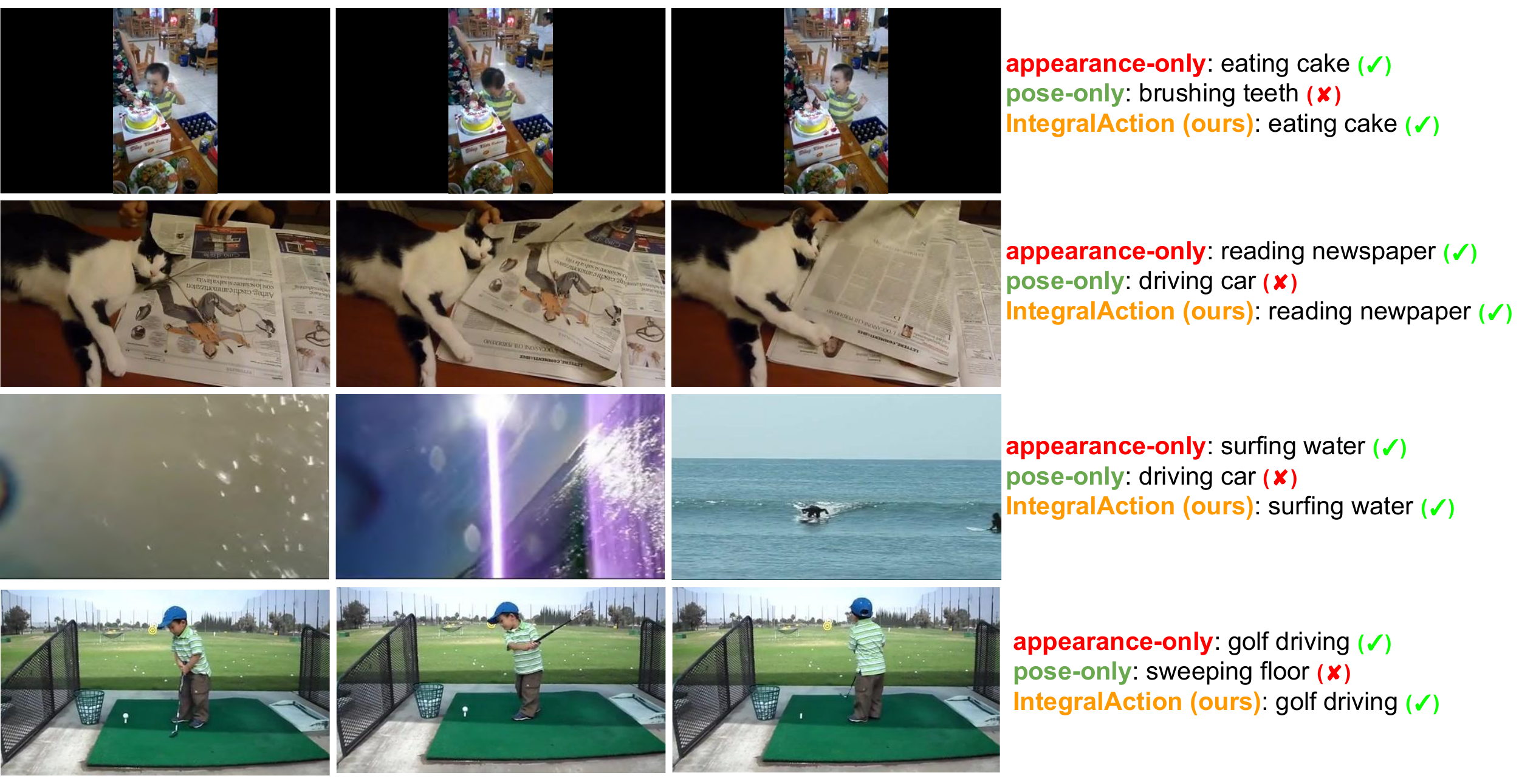}
\end{center}
\vspace*{-3mm}
   \caption{
Qualitative results of appearance-only, pose-only, and the proposed IntegralAction.
   }
\vspace*{-3mm}
\label{fig:rgb_qualitative}
\end{figure*}

\begin{figure*}
\begin{center}
\includegraphics[width=1.0\linewidth]{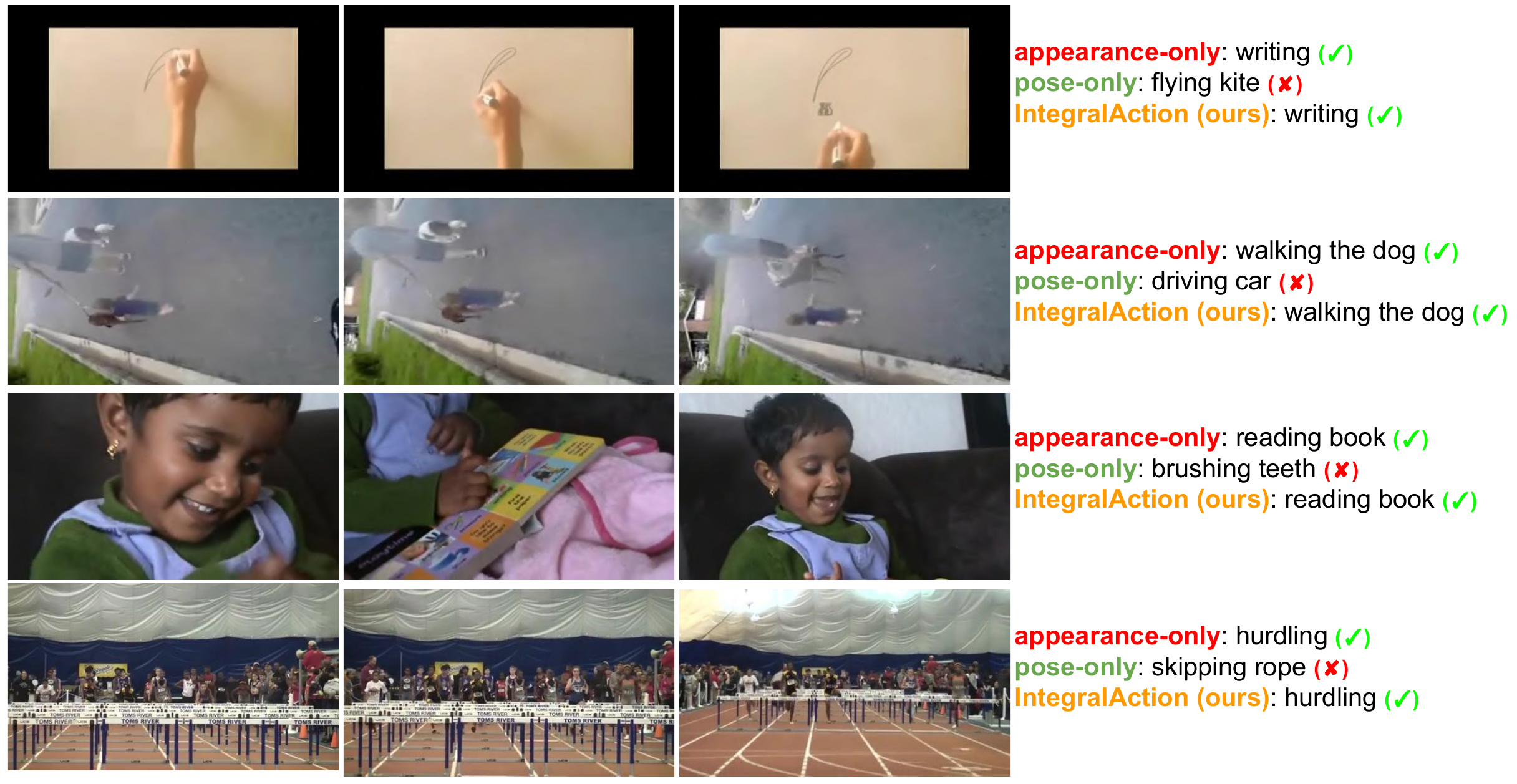}
\end{center}
\vspace*{-3mm}
   \caption{
Qualitative results of appearance-only, pose-only, and the proposed IntegralAction.
   }
\vspace*{-3mm}
\label{fig:rgb_qualitative2}
\end{figure*}

\begin{figure*}
\begin{center}
\includegraphics[width=1.0\linewidth]{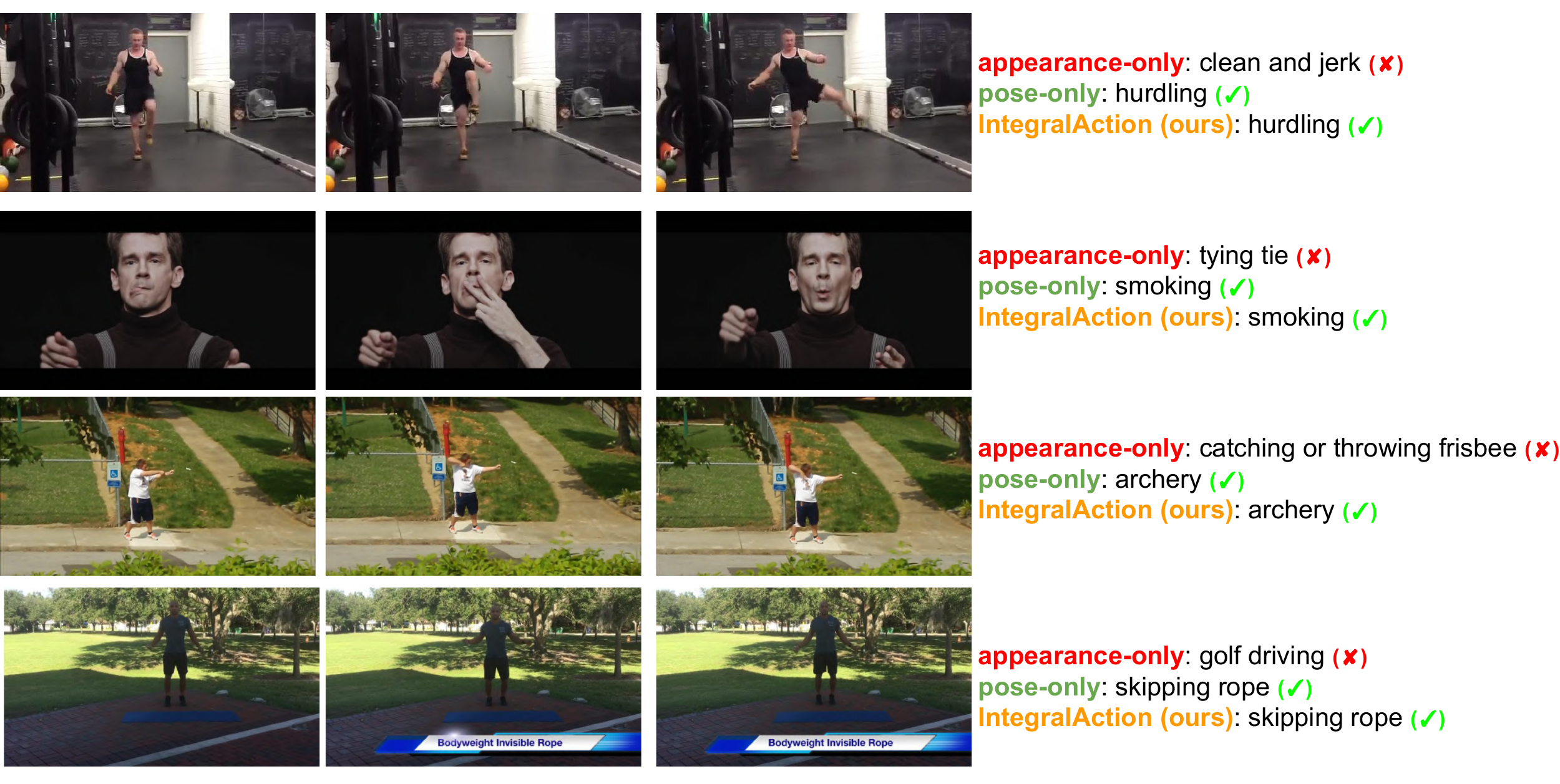}
\end{center}
\vspace*{-3mm}
   \caption{
Qualitative results of appearance-only, pose-only, and the proposed IntegralAction.
   }
\vspace*{-3mm}
\label{fig:pose_qualitative}
\end{figure*}

\begin{figure*}
\begin{center}
\includegraphics[width=1.0\linewidth]{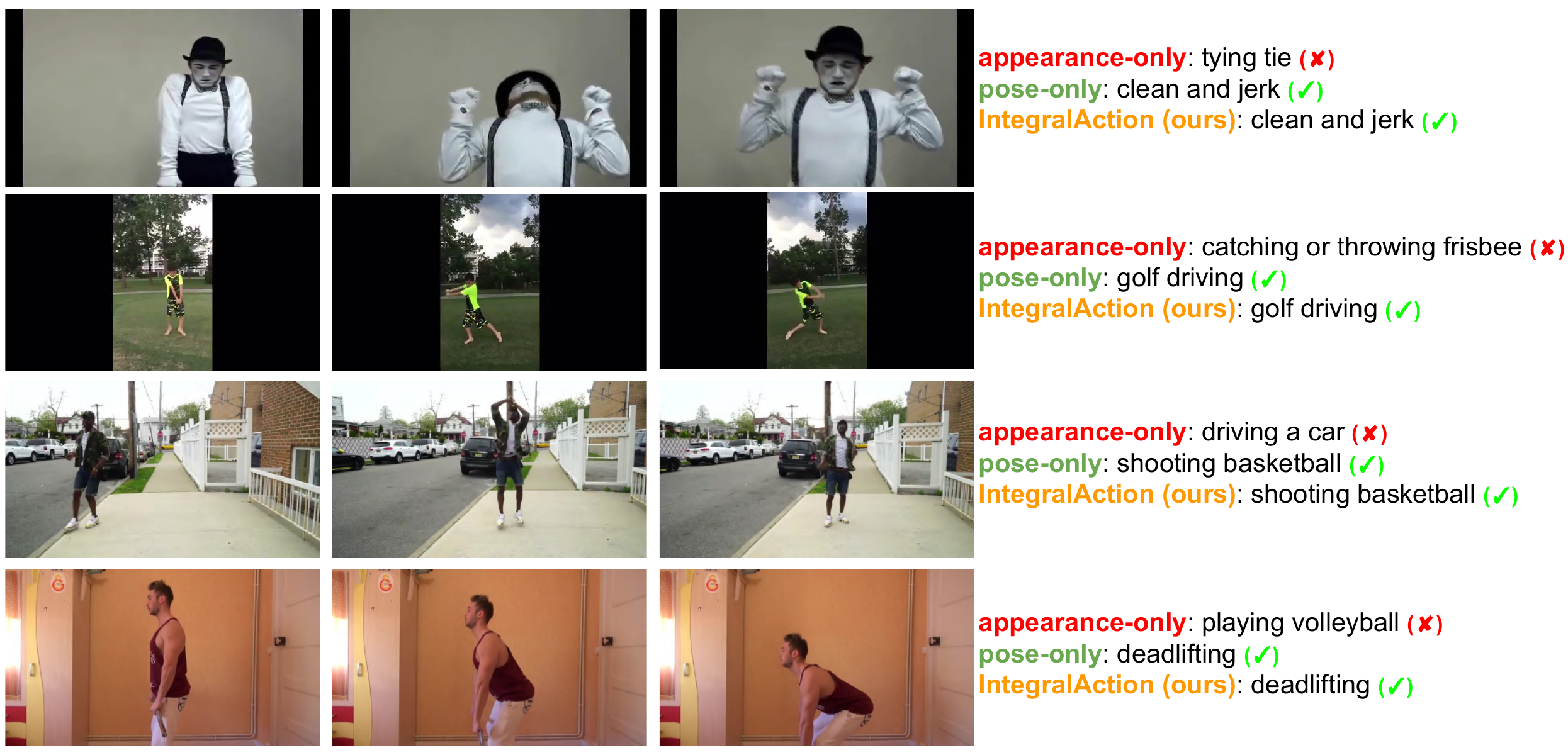}
\end{center}
\vspace*{-3mm}
   \caption{
Qualitative results of appearance-only, pose-only, and the proposed IntegralAction.
   }
\vspace*{-3mm}
\label{fig:pose_qualitative2}
\end{figure*}

\begin{figure*}
\begin{center}
\includegraphics[width=1.0\linewidth]{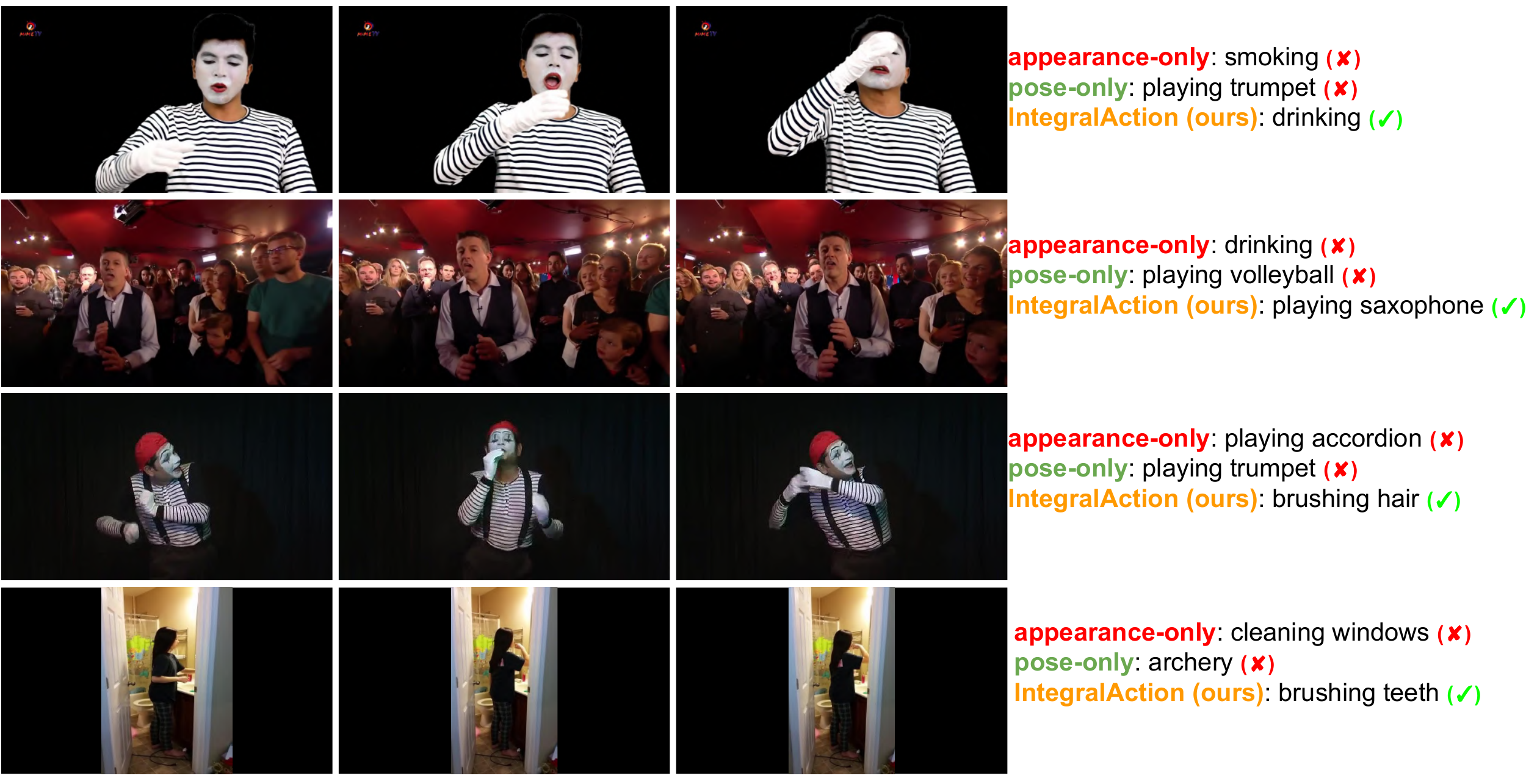}
\end{center}
\vspace*{-3mm}
   \caption{
Qualitative results of appearance-only, pose-only, and the proposed IntegralAction.
   }
\vspace*{-3mm}
\label{fig:both_qualitative}
\end{figure*}

Figures~\ref{fig:rgb_qualitative}, ~\ref{fig:rgb_qualitative2}, ~\ref{fig:pose_qualitative}, ~\ref{fig:pose_qualitative2}, and ~\ref{fig:both_qualitative} show qualitative results from the appearance-only, pose-only, and the proposed IntegralAction.
Interestingly, our IntegralAction often succeeds in recognizing correct actions even when both the appearance-only and pose-only fail and so does the oracle selection, as shown in Fig.~\ref{fig:both_qualitative}.
We found that this happens when the appearance-only model is fooled by focusing on the contextual information such as background scene and objects, and the pose-only model suffers from the context ambiguity because the input pose sequence can be mapped to multiple action classes.
For example, the second example of Fig.~\ref{fig:both_qualitative} shows that the appearance-only model is fooled by drinking people, and the pose-only model suffers from the context ambiguity.
As the input pose sequence does not contain finger keypoints, the pose-only model predicts the input pose is about playing volleyball based on the given body keypoints.
The input pose sequence may need to contain richer geometric information of the human body for better performance, for example, finger keypoints and finally, a 3D mesh of the human.
Also, improving the integration part to more effectively combine the context from the appearance stream and the human motion from the pose stream should also be studied.

\section{Network architecture of IntegralAction}
In this section, we provide the detailed network architectures used in our paper. 
Table~\ref{tab:architecture1} shows the network architecture we used in Section 4.2 of the main manuscript, while Table~\ref{tab:architecture2} shows the network architecture we used in Section 4.3 and 4.4 of the main manuscript.

\begin{table*}
\footnotesize
\centering
\setlength\tabcolsep{1.0pt}
\def\arraystretch{1.1}
\caption{The top-1 accuracy for each action comparison between appearance-only, pose-only, our IntegralAction, and the oracle selection on Kinetics50.}
\label{table:label_accuracy_compare_kinetics50}
\begin{tabular}{C{5.5cm}|C{3.0cm}C{2.0cm}C{3.5cm}|C{3.0cm}}
\specialrule{.1em}{.05em}{.05em}
classes & appearance-only & pose-only & \textbf{IntegralAction (ours)} & oracle selection \\ \hline
surfing water & 87.5 & 25.0 & \textbf{91.7} & 87.5 \\
shooting goal (soccer) & \textbf{51.0} & 10.2 & 49.0 & 51.0 \\
hitting baseball & 84.0 & 54.0 & \textbf{88.0} & 88.0 \\
playing bass guitar & \textbf{86.0} & 44.0 & 78.0 & 88.0 \\
reading book & \textbf{66.0} & 38.0 & 60.0 & 70.0 \\
juggling soccer ball & 58.0 & 66.0 & \textbf{70.0} & 80.0 \\
dribbling basketball & \textbf{74.0} & 58.0 & 72.0 & 82.0 \\
playing accordion & \textbf{91.8} & 75.5 & 89.8 & 91.8 \\
catching or throwing baseball & 39.6 & 10.4 & \textbf{41.7} & 50.0 \\
archery & \textbf{77.6} & 30.6 & 75.5 & 79.6 \\
tying tie & \textbf{86.0} & 44.0 & \textbf{86.0} & 88.0 \\
skiing (not slalom or crosscountry) & \textbf{95.9} & 63.3 & \textbf{95.9} & 95.9 \\
brushing hair & \textbf{62.0} & 34.0 & 60.0 & 68.0 \\
hurdling & \textbf{92.0} & 66.0 & 90.0 & 94.0 \\
playing violin & \textbf{76.0} & 60.0 & 74.0 & 86.0 \\
playing volleyball & \textbf{79.2} & 45.8 & 77.1 & 81.2 \\
deadlifting & 87.8 & 87.8 & \textbf{91.8} & 93.9 \\
skipping rope & 67.3 & 77.6 & \textbf{85.7} & 85.7 \\
playing piano & \textbf{78.0} & 38.0 & 76.0 & 80.0 \\
writing & \textbf{72.0} & 22.0 & \textbf{72.0} & 74.0 \\
climbing a rope & 78.0 & \textbf{82.0} & 78.0 & 88.0 \\
dunking basketball & 56.2 & 41.7 & \textbf{60.4} & 68.8 \\
playing basketball & \textbf{58.0} & 32.0 & \textbf{58.0} & 66.0 \\
brushing teeth & 66.0 & 44.0 & \textbf{68.0} & 72.0 \\
drinking & 30.6 & 14.3 & \textbf{32.7} & 36.7 \\
driving car & \textbf{91.7} & 39.6 & 87.5 & 91.7 \\
walking the dog & \textbf{93.9} & 65.3 & 91.8 & 95.9 \\
playing saxophone & \textbf{80.0} & 54.0 & 78.0 & 86.0 \\
playing trumpet & \textbf{83.7} & 57.1 & \textbf{83.7} & 85.7 \\
bowling & \textbf{93.9} & 38.8 & 87.8 & 93.9 \\
punching person (boxing) & \textbf{79.2} & 60.4 & 75.0 & 83.3 \\
cleaning windows & \textbf{76.0} & 16.0 & \textbf{76.0} & 80.0 \\
clean and jerk & \textbf{91.8} & 87.8 & \textbf{91.8} & 93.9 \\
eating cake & \textbf{58.0} & 22.0 & 48.0 & 64.0 \\
flying kite & \textbf{90.0} & 54.0 & \textbf{90.0} & 96.0 \\
opening bottle & 52.0 & 18.0 & \textbf{70.0} & 58.0 \\
canoeing or kayaking & \textbf{94.0} & 38.0 & 90.0 & 94.0 \\
reading newspaper & \textbf{54.0} & 8.0 & 38.0 & 54.0 \\
skiing slalom & \textbf{86.0} & 76.0 & \textbf{86.0} & 92.0 \\
playing guitar & \textbf{80.0} & 56.0 & 76.0 & 84.0 \\
eating ice cream & \textbf{54.0} & 20.0 & 46.0 & 66.0 \\
climbing ladder & 68.0 & 40.0 & \textbf{72.0} & 74.0 \\
juggling balls & 81.6 & 79.6 & \textbf{85.7} & 91.8 \\
shooting basketball & \textbf{30.6} & 16.3 & 22.4 & 40.8 \\
catching or throwing frisbee & \textbf{56.0} & 8.0 & 48.0 & 58.0 \\
sweeping floor & \textbf{72.0} & 42.0 & \textbf{72.0} & 76.0 \\
playing tennis & \textbf{93.9} & 69.4 & \textbf{93.9} & 98.0 \\
sword fighting & 38.8 & 32.7 & \textbf{40.8} & 53.1 \\
smoking & \textbf{55.1} & 36.7 & 53.1 & 65.3 \\
golf driving & 84.0 & 80.0 & \textbf{86.0} & 86.0 \\ \hline
average & \textbf{72.8} & 45.6 & 72.2 & 78.2 \\
 \specialrule{.1em}{.05em}{.05em}
\end{tabular}
\end{table*}

\begin{table*}
\footnotesize
\centering
\setlength\tabcolsep{1.0pt}
\def\arraystretch{1.1}
\caption{The top-1 accuracy for each action comparison between appearance-only, pose-only, our IntegralAction, and the oracle selection on Mimetics.}
\label{table:label_accuracy_compare_mimetics}
\begin{tabular}{C{5.5cm}|C{3.0cm}C{2.0cm}C{3.5cm}|C{3.0cm}}
\specialrule{.1em}{.05em}{.05em}
classes & appearance-only & pose-only & \textbf{IntegralAction (ours)} & oracle selection \\ \hline
surfing water & 0.0 & 0.0 & \textbf{12.5} & 0.0 \\
shooting goal (soccer) & \textbf{18.2} & 9.1 & \textbf{18.2} & 27.3 \\
hitting baseball & 0.0 & \textbf{10.0} & \textbf{10.0} & 10.0 \\
playing bass guitar & 9.1 & 18.2 & \textbf{27.3} & 27.3 \\
reading book & \textbf{11.1} & \textbf{11.1} & 0.0 & 22.2 \\
juggling soccer ball & 16.7 & \textbf{41.7} & 33.3 & 41.7 \\
dribbling basketball & 0.0 & \textbf{61.5} & 23.1 & 61.5 \\
playing accordion & 10.0 & 40.0 & \textbf{50.0} & 50.0 \\
catching or throwing baseball & \textbf{14.3} & \textbf{14.3} & 0.0 & 28.6 \\
archery & 6.7 & \textbf{33.3} & 20.0 & 33.3 \\
tying tie & \textbf{16.7} & \textbf{16.7} & \textbf{16.7} & 16.7 \\
skiing (not slalom or crosscountry) & \textbf{12.5} & \textbf{12.5} & \textbf{12.5} & 12.5 \\
brushing hair & 11.8 & 41.2 & \textbf{47.1} & 41.2 \\
hurdling & 11.1 & \textbf{55.6} & 22.2 & 55.6 \\
playing violin & 11.8 & 17.6 & \textbf{23.5} & 17.6 \\
playing volleyball & 23.1 & 23.1 & \textbf{38.5} & 38.5 \\
deadlifting & 11.1 & \textbf{100.0} & \textbf{100.0} & 100.0 \\
skipping rope & 25.0 & \textbf{83.3} & 75.0 & 83.3 \\
playing piano & \textbf{11.8} & \textbf{11.8} & \textbf{11.8} & 17.6 \\
writing & \textbf{0.0} & \textbf{0.0} & \textbf{0.0} & 0.0 \\
climbing a rope & 0.0 & 42.9 & \textbf{50.0} & 42.9 \\
dunking basketball & 0.0 & \textbf{33.3} & \textbf{33.3} & 33.3 \\
playing basketball & 23.1 & 23.1 & \textbf{30.8} & 30.8 \\
brushing teeth & 35.7 & 35.7 & \textbf{42.9} & 50.0 \\
drinking & 10.0 & 20.0 & \textbf{30.0} & 25.0 \\
driving car & \textbf{0.0} & \textbf{0.0} & \textbf{0.0} & 0.0 \\
walking the dog & \textbf{7.7} & \textbf{7.7} & \textbf{7.7} & 7.7 \\
playing saxophone & 0.0 & 7.7 & \textbf{15.4} & 7.7 \\
playing trumpet & 0.0 & \textbf{50.0} & \textbf{50.0} & 50.0 \\
bowling & \textbf{8.3} & 0.0 & \textbf{8.3} & 8.3 \\
punching person (boxing) & 9.1 & 36.4 & \textbf{45.5} & 36.4 \\
cleaning windows & \textbf{6.7} & 0.0 & 0.0 & 6.7 \\
clean and jerk & 15.4 & \textbf{92.3} & \textbf{92.3} & 92.3 \\
eating cake & 0.0 & \textbf{11.8} & 0.0 & 11.8 \\
flying kite & 10.0 & 0.0 & \textbf{20.0} & 10.0 \\
opening bottle & \textbf{0.0} & \textbf{0.0} & \textbf{0.0} & 0.0 \\
canoeing or kayaking & 0.0 & \textbf{21.4} & 14.3 & 21.4 \\
reading newspaper & 0.0 & \textbf{11.1} & 0.0 & 11.1 \\
skiing slalom & 0.0 & \textbf{10.0} & \textbf{10.0} & 10.0 \\
playing guitar & 7.1 & \textbf{14.3} & \textbf{14.3} & 14.3 \\
eating ice cream & \textbf{0.0} & \textbf{0.0} & \textbf{0.0} & 0.0 \\
climbing ladder & 7.7 & \textbf{15.4} & 7.7 & 23.1 \\
juggling balls & 42.9 & \textbf{57.1} & \textbf{57.1} & 71.4 \\
shooting basketball & 8.3 & 8.3 & \textbf{16.7} & 16.7 \\
catching or throwing frisbee & \textbf{50.0} & 0.0 & 40.0 & 50.0 \\
sweeping floor & \textbf{0.0} & \textbf{0.0} & \textbf{0.0} & 0.0 \\
playing tennis & 5.6 & \textbf{16.7} & 11.1 & 22.2 \\
sword fighting & 46.7 & \textbf{66.7} & \textbf{66.7} & 73.3 \\
smoking & 26.7 & 20.0 & \textbf{33.3} & 33.3 \\
golf driving & 14.3 & \textbf{64.3} & 50.0 & 64.3 \\ \hline
average & 11.2 & 26.0 & \textbf{26.5} & 30.7 \\
 \specialrule{.1em}{.05em}{.05em}
\end{tabular}
\vspace*{-7mm}
\end{table*}

\begin{table*}
\scriptsize
\centering
\setlength\tabcolsep{1.0pt}
\def\arraystretch{1.1}
\caption{The top-1 accuracy for each action comparison between appearance-only, pose-only, our IntegralAction, and the oracle selection on NTU-RGBD.}
\label{table:label_accuracy_compare_ntu}
\begin{tabular}{C{5.5cm}|C{3.0cm}C{2.0cm}C{3.5cm}|C{3.0cm}}
\specialrule{.1em}{.05em}{.05em}
classes & appearance-only & pose-only & \textbf{IntegralAction (ours)} & oracle selection \\ \hline
drink water & 90.9 & 82.8 & \textbf{92.0} & 96.4 \\
eat meal & 80.0 & 66.9 & \textbf{82.5} & 82.9 \\
brush teeth & \textbf{92.3} & 80.1 & 91.5 & 95.6 \\
brush hair & 95.2 & 87.5 & \textbf{96.7} & 97.4 \\
drop & \textbf{97.1} & 78.5 & \textbf{97.1} & 98.5 \\
pick up & 96.7 & 93.8 & \textbf{98.2} & 98.2 \\
throw & \textbf{90.2} & 83.3 & \textbf{90.2} & 94.2 \\
sit down & 95.6 & 96.0 & \textbf{98.5} & 98.9 \\
stand up & 98.5 & 96.7 & \textbf{99.3} & 99.6 \\
clapping & \textbf{79.5} & 64.8 & 78.8 & 89.4 \\
reading & \textbf{73.9} & 48.2 & 70.6 & 82.0 \\
writing & 64.3 & 40.4 & \textbf{64.7} & 77.2 \\
tear up paper & 93.0 & 80.8 & \textbf{93.7} & 95.9 \\
put on jacket & \textbf{100.0} & 97.8 & 99.6 & 100.0 \\
take off jacket & \textbf{98.2} & 94.6 & 96.4 & 98.9 \\
put on a shoe & \textbf{90.8} & 58.6 & 83.2 & 92.7 \\
take off a shoe & \textbf{82.8} & 55.8 & 75.5 & 89.8 \\
put on glasses & 85.7 & 85.3 & \textbf{92.6} & 96.0 \\
take off glasses & 89.8 & 86.1 & \textbf{92.3} & 93.4 \\
put on a hat/cap & \textbf{99.6} & 92.3 & 98.2 & 100.0 \\
take off a hat/cap & 98.2 & 94.1 & \textbf{98.5} & 99.3 \\
cheer up & \textbf{94.5} & 90.9 & 93.8 & 96.7 \\
hand waving & 87.2 & 87.2 & \textbf{90.9} & 93.4 \\
kicking something & \textbf{98.2} & 93.1 & \textbf{98.2} & 99.3 \\
reach into pocket & \textbf{81.8} & 72.6 & 81.4 & 85.8 \\
hopping & 94.9 & \textbf{96.4} & \textbf{96.4} & 96.7 \\
jump up & \textbf{100.0} & 99.6 & \textbf{100.0} & 100.0 \\
phone call & \textbf{89.1} & 78.9 & 78.2 & 94.9 \\
play with phone/tablet & 70.2 & 58.2 & \textbf{76.4} & 81.5 \\
type on a keyboard & 88.4 & 66.5 & \textbf{89.8} & 93.8 \\
point to something & 88.8 & 75.7 & \textbf{90.6} & 92.4 \\
taking a selfie & 88.0 & 85.5 & \textbf{93.1} & 94.6 \\
check time (from watch) & 88.4 & 84.8 & \textbf{92.8} & 97.1 \\
rub two hands & 72.5 & 75.4 & \textbf{79.7} & 91.7 \\
nod head/bow & 93.8 & 90.2 & \textbf{95.7} & 97.1 \\
shake head & 94.9 & 83.2 & \textbf{98.2} & 97.8 \\
wipe face & 83.7 & 76.4 & \textbf{91.7} & 95.7 \\
salute & \textbf{95.7} & 90.6 & \textbf{95.7} & 97.8 \\
put palms together & 82.2 & 88.8 & \textbf{92.4} & 95.7 \\
cross hands in front & 96.0 & 94.9 & \textbf{97.5} & 97.8 \\
sneeze/cough & 68.5 & 69.6 & \textbf{77.2} & 80.4 \\
staggering & 98.6 & 96.7 & \textbf{98.9} & 99.6 \\
falling down & 98.2 & 96.0 & \textbf{98.9} & 98.5 \\
headache & 68.5 & 68.1 & \textbf{75.0} & 84.1 \\
chest pain & 88.0 & 88.0 & \textbf{92.0} & 96.4 \\
back pain & 95.3 & 85.5 & \textbf{97.1} & 98.9 \\
neck pain & 83.7 & 77.2 & \textbf{89.1} & 92.8 \\
nausea/vomiting & 87.6 & 83.6 & \textbf{90.5} & 93.1 \\
fan self & 86.5 & 86.9 & \textbf{88.4} & 96.0 \\
punch/slap & 88.3 & 89.1 & \textbf{92.0} & 95.3 \\
kicking & \textbf{98.6} & 91.7 & 97.8 & 99.3 \\
pushing & 97.5 & 94.9 & \textbf{98.5} & 99.6 \\
pat on back & 97.8 & 85.5 & \textbf{98.2} & 99.3 \\
point finger & 97.5 & 89.1 & \textbf{97.8} & 98.6 \\
hugging & 99.3 & 97.4 & \textbf{99.6} & 100.0 \\
giving object & \textbf{95.3} & 87.3 & 94.2 & 97.1 \\
touch pocket & \textbf{96.7} & 92.0 & 96.0 & 99.3 \\
shaking hands & 98.2 & 94.6 & \textbf{100.0} & 99.6 \\
walking towards & \textbf{100.0} & 98.9 & \textbf{100.0} & 100.0 \\
walking apart & \textbf{100.0} & 97.1 & 98.9 & 100.0 \\ \hline
average & 90.4 & 83.7 & \textbf{91.7} & 95.1 \\
 \specialrule{.1em}{.05em}{.05em}
\end{tabular}
\end{table*}

\clearpage

\clearpage

\begin{table*}
  \begin{center}
    \caption{The network architecture details of IntegralAction in Section 4.2 of the main manuscript.
    The dimensions of kernels are denoted by (T$\times$S$^2$, C) for the temporal, spatial, and channel sizes.
    The strides and output size are denoted by (T$\times$S$^2$) for the temporal and spatial sizes.
}
    \label{tab:architecture1}     
    \begin{tabular}{c|c|c|c}     
 \specialrule{.1em}{.05em}{.05em}    
      layers & appearance stream & pose stream & output size \\
 \hline 
      \multirow{2}{*}{input} & \multirow{2}{*}{RGB frames} & \multirow{2}{*}{keypoint heatmaps+PAFs} & appearance: 8$\times$224$^2$  \\ 
      & & &  pose: 32$\times$56$^2$\\
      \hline
      \multirow{2}{*}{conv$_1$} & \multirow{2}{*}{1$\times$7$^2$, 64, stride 1$\times$2$^2$} & \multirow{2}{*}{1$\times$3$^2$, 64} & appearance: 8$\times$112$^2$ \\
      & & & pose: 32$\times$56$^2$ \\
  \hline
      \multirow{6}{*}{res$_2$}& 1$\times$3$^2$ max pool, stride 1$\times$2$^2$ &  \multirow{6}{*}{$\begin{bmatrix}  $TSM$  \\    $1$\times$3$^2$, 64$ \\$1$\times$3$^2$, 64$\end{bmatrix}\times$2}  & \multirow{5}{*}{appearance: 8$\times$56$^2$} \\  
      \cline{2-2}  
          & & & \\[-0.7em]
      & $\begin{bmatrix}  $TSM$ \\ $1$\times$1$^2$, 256$ \\    $1$\times$3$^2$, 256$  \\   $1$\times$1$^2$, 256$\end{bmatrix}\times$3 & & pose: 32$\times$56$^2$ \\ 
          & & & \\[-0.7em]
    \hline
    & & & \multirowcell{5}{appearance: 8$\times$28$^2$ \\ pose: 32$\times$28$^2$ }\\[-0.7em]
  res$_3$ & $\begin{bmatrix} $TSM$\\ $1$\times$1$^2$, 512$ \\    $1$\times$3$^2$, 512$ \\ $1$\times$1$^2$, 512$\end{bmatrix}\times$4
  & $\begin{bmatrix} $TSM$\\$1$\times$3$^2$, 128$ \\    $1$\times$3$^2$, 128$ \end{bmatrix}\times$2  
  &   \\
      & & & \\[-0.7em]
  \hline  
    & & & \multirowcell{5}{appearance: 8$\times$14$^2$ \\ pose: 32$\times$14$^2$ } \\[-0.7em]  
  res$_4$ & $\begin{bmatrix} $TSM$\\ $1$\times$1$^2$, 1024$ \\    $1$\times$3$^2$, 1024$ \\ $1$\times$1$^2$, 1024$\end{bmatrix}\times$6
  & $\begin{bmatrix} $TSM$\\$1$\times$3$^2$, 256$ \\    $1$\times$3$^2$, 256$ \end{bmatrix}\times$2  
  &  \\
      & & & \\[-0.7em]
  \hline  
    & & & \multirowcell{5}{appearance: 8$\times$7$^2$ \\ pose: 32$\times$7$^2$ } \\[-0.7em]  
    res$_5$ & $\begin{bmatrix} $TSM$\\ $1$\times$1$^2$, 2048$ \\    $1$\times$3$^2$, 2048$ \\ $1$\times$1$^2$, 2048$ \end{bmatrix}\times$3
    & $\begin{bmatrix} $TSM$\\$1$\times$3$^2$, 512$ \\    $1$\times$3$^2$, 512$ \end{bmatrix}\times$2  
    &  \\
        & & & \\[-0.7em]
  \hline  
 \multirow{2}{*}{pool} & \multirow{2}{*}{global average pool} & \multirow{2}{*}{global average pool} & appearance: 8$\times$1$^2$\\ 
 & & & pose: 32$\times$1$^2$ \\
  \hline
  \multirowcell{3}{feature align \\ (TCB$_\text A$,TCB$_\text P$)} & \multirowcell{3}{1$\times$1$^2$, 512 \\ layer normalization} & 4$\times$1$^2$ avg pool, stride 4$\times$1$^2$ & \multirow{6}{*}{both: 8$\times$1$^2$}\\ 
  \cline{3-3}
  & & 1$\times$1$^2$, 512  \\
  &  & layer normalization & \\  
  \cline{1-3}
  \multirowcell{2}{pose-driven\\gating (CGB)} &\multirow{2}{*}{($1-\mathbf{G}$) element-wise product} & ($\mathbf{G}:$1$\times$1$^2$, 512) & \\
   & & $\mathbf{G}$ element-wise product & \\
  \cline{1-3}
  aggregation & \multicolumn{2}{c|}{element-wise addition} & \\   
  \hline 
  classifier & \multicolumn{2}{c|}{fully-connected layer} & \# of classes\\
 \specialrule{.1em}{.05em}{.05em}
    \end{tabular}
  \end{center}
\end{table*}

\begin{table*}
  \begin{center}
    \caption{The network architecture details of IntegralAction in Section 4.3 and 4.4 of the main manuscript.
        The dimensions of kernels are denoted by (T$\times$S$^2$, C) for the temporal, spatial, and channel sizes.
    The strides and output size are denoted by (T$\times$S$^2$) for the temporal and spatial sizes.
    }
    \label{tab:architecture2}    
    \begin{tabular}{c|c|c|c}     
 \specialrule{.1em}{.05em}{.05em}    
      layers & appearance stream & pose stream & output size \\
 \hline 
      \multirow{2}{*}{input} & \multirow{2}{*}{RGB frames} & \multirow{2}{*}{keypoint heatmaps+PAFs} & appearance: 8$\times$224$^2$  \\ 
      & &  &  pose: 8$\times$56$^2$\\
      \hline
      \multirow{2}{*}{conv$_1$} & \multirow{2}{*}{1$\times$7$^2$, 64, stride 1$\times$2$^2$} & \multirow{2}{*}{1$\times$3$^2$, 64} & appearance: 8$\times$112$^2$ \\
      & & & pose: 8$\times$56$^2$ \\
  \hline
      \multirow{5}{*}{res$_2$}& 1$\times$3$^2$ max pool, stride 1$\times$2$^2$ &  \multirow{5}{*}{$\begin{bmatrix}  $TSM$ \\$1$\times$3$^2$, 64$ \\    $1$\times$3$^2$, 64$ \end{bmatrix}\times$2}  & \multirow{5}{*}{both: 8$\times$56$^2$ } \\  
      \cline{2-2}  
                & & & \\[-0.7em]
      & $\begin{bmatrix}  $TSM$ \\ $1$\times$3$^2$, 64$ \\    $1$\times$3$^2$, 64$ \end{bmatrix}\times$2 & & \\ 
                & & & \\[-0.7em]
    \hline
              & & & \\[-0.7em]
  res$_3$ & $\begin{bmatrix} $TSM$ \\ $1$\times$3$^2$, 128$ \\    $1$\times$3$^2$, 128$ \end{bmatrix}\times$2 
  & $\begin{bmatrix} $TSM$\\$1$\times$3$^2$, 128$ \\    $1$\times$3$^2$, 128$ \end{bmatrix}\times$2 
  & both: 8$\times$28$^2$   \\
            & & & \\[-0.7em]
  \hline  
            & & & \\[-0.7em]
  res$_4$ & $\begin{bmatrix} $TSM$\\ $1$\times$3$^2$, 256$ \\    $1$\times$3$^2$, 256$ \end{bmatrix}\times$2 
  & $\begin{bmatrix} $TSM$\\$1$\times$3$^2$, 256$ \\    $1$\times$3$^2$, 256$ \end{bmatrix}\times$2  
  & both: 8$\times$14$^2$   \\
            & & & \\[-0.7em]
  \hline  
            & & & \\[-0.7em]
    res$_5$ & $\begin{bmatrix} $TSM$\\ $1$\times$3$^2$, 512$ \\    $1$\times$3$^2$, 512$ \end{bmatrix}\times$2 
    & $\begin{bmatrix} $TSM$\\$1$\times$3$^2$, 512$ \\    $1$\times$3$^2$, 512$ \end{bmatrix}\times$2  
    & both: 8$\times$7$^2$   \\
              & & & \\[-0.7em]
  \hline  
 pool & global average pool & global average pool & \multirow{6}{*}{both: 8$\times$1$^2$}\\ 
      \cline{1-3}
  feature align & 1$\times$1$^2$, 512 & 1$\times$1$^2$, 512 & \\
  (TCB$_\text A$,TCB$_\text P$) & layer normalization & layer normalization & \\  
  \cline{1-3}
   pose-driven &\multirow{2}{*}{($1-\mathbf{G}$) element-wise product} & ($\mathbf{G}:$1$\times$1$\times$1, 512) & \\
   gating (CGB) & & $\mathbf{G}$ element-wise product & \\
  \cline{1-3}
  aggregation & \multicolumn{2}{c|}{element-wise addition} & \\   
  \hline 
  classifier & \multicolumn{2}{c|}{fully-connected layer} & \# of classes\\ 
 \specialrule{.1em}{.05em}{.05em}
    \end{tabular}
  \end{center}
\end{table*}

\clearpage

{\small
\bibliographystyle{ieee_fullname}
\bibliography{bib/bib}
}

{\small
\bibliographystyle{ieee_fullname}
\bibliography{main}
}

\end{document}